\documentclass[10pt,twocolumn,letterpaper]{article}

\usepackage[pagenumbers]{cvpr} 

\usepackage{graphicx}
\usepackage{amsmath}
\usepackage{amssymb}
\usepackage{booktabs}
\usepackage{titling}

\usepackage{amsmath,amsfonts,bm}

\def\eqref#1{equation~\ref{#1}}

\def\1{\bm{1}}

\DeclareMathAlphabet{\mathsfit}{\encodingdefault}{\sfdefault}{m}{sl}
\SetMathAlphabet{\mathsfit}{bold}{\encodingdefault}{\sfdefault}{bx}{n}

\usepackage{url}
\usepackage{color,soul}
\usepackage{multirow,graphicx,paralist,bigdelim}
\usepackage{colortbl}
\usepackage{tabularx}
\usepackage{enumitem}
\usepackage{pifont}
\usepackage{bbm}
\usepackage{mathtools}

\makeatletter
\@namedef{ver@everyshi.sty}{}
\makeatother
\usepackage{tikz}
\usepackage{pgfplots}
\pgfplotsset{compat=default}
\usetikzlibrary{pgfplots.groupplots}
\usepackage{tabu}
\usepackage[pagebackref,breaklinks,colorlinks]{hyperref}
\usepackage{pifont}
\usepackage{listings}
\usepackage[capitalize]{cleveref}
\crefname{section}{Sec.}{Secs.}
\Crefname{section}{Section}{Sections}
\Crefname{table}{Table}{Tables}
\crefname{table}{Tab.}{Tabs.}

\def\cvprPaperID{6082} 
\def\confName{CVPR}
\def\confYear{2022}

\begin{document}

\vspace{-2.0cm}
\date{}
\title{\Large
\textbf{A Deeper Dive Into What Deep Spatiotemporal Networks Encode: \\
Quantifying Static vs. Dynamic Information}}
\author{Matthew Kowal$^{1,2}$,
Mennatullah Siam$^{1}$, 
Md Amirul Islam$^{2,3}$ \\
Neil D. B. Bruce$^{2,5}$,
Richard P. Wildes$^{1,4}$,
Konstantinos G. Derpanis$^{1,2,4}$ \\
{\small $^{1}$York University, $^{2}$Vector Institute for AI, $^{3}$Ryerson University, $^{4}$Samsung AI Centre Toronto, $^{5}$University of Guelph}
\\
{\small 
\texttt{\{m2kowal,msiam,wildes,kosta\}@eecs.yorku.ca}, \texttt{mdamirul@ryerson.ca}, \texttt{brucen@uoguelph.ca}}
}
\maketitle

\begin{abstract}
Deep spatiotemporal models are used in a variety of computer vision tasks, such as action recognition and video object segmentation. Currently, there is a limited understanding of what information is captured by these models in their intermediate representations. For example, while it has been observed that action recognition algorithms are heavily influenced by visual appearance in single static frames, 
there is no quantitative methodology for evaluating such static bias in the latent representation compared to bias toward dynamic information (e.g.\ motion).
We tackle this challenge by proposing a novel approach for quantifying the static and dynamic biases of any spatiotemporal model. To show the efficacy of our approach, we analyse two widely studied tasks, action recognition and video object segmentation. 
Our key findings are threefold: (i) Most examined spatiotemporal models are biased toward static information; although, certain two-stream architectures with cross-connections show a better balance between the static and dynamic information captured. (ii) Some datasets that are commonly assumed to be biased toward dynamics are actually biased toward static information. (iii) Individual units (channels) in an architecture can be biased toward static, dynamic or a combination of the two.~\footnote{ \href{https://yorkucvil.github.io/Static-Dynamic-Interpretability/}{Project page and code}}
\vspace{-1.0cm}
\end{abstract}

\section{Introduction}
\begin{figure}[t]
\centering
    \resizebox{0.48\textwidth}{!}{
    \includegraphics{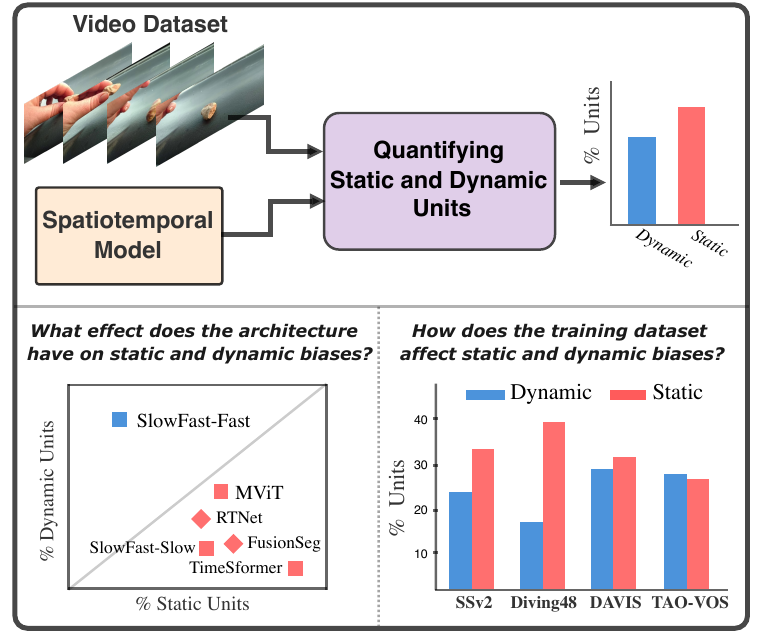}}
    \caption{We introduce a general technique that, given a model and a video dataset, can quantify the bias in any intermediate representation within the model toward encoding static (red) or dynamic (blue) information. We use this technique to study the tasks of action recognition (squares) and video object segmentation (diamonds) and explore the effect of architectures and training datasets on static and dynamic biases.}
\vspace{-2em}
\end{figure}\label{fig:motivation}


\vspace{-0.1cm}
This paper focuses on the problem of interpreting the information learned by deep neural networks (DNNs) trained for video understanding tasks. Interpreting deep spatiotemporal models is a largely understudied topic in computer vision despite their achieving state-of-the-art performance on video understanding tasks, such as action recognition~\cite{zhu2020comprehensive} and video object segmentation~\cite{wang2021survey}. These models are trained in an end-to-end fashion to learn discriminative static and dynamic features over space and time. 
Here, we use the term \textit{static} to refer to attributes that can be extracted from a single image (\eg color and texture) and the term \textit{dynamic} to attributes that arise from consideration of multiple frames (\eg motion and dynamic texture). 

While this learning-based paradigm has led to great success across a wide range of tasks, the internal representations of the learned models remain largely opaque. This lack of explainability is unsatisfying from both scientific and application perspectives. From a scientific perspective, there is limited understanding of what information is driving the decision-making underlying the network output. Elucidating
the decision-making process may yield directions to improve models. From an applications perspective, there have been multiple cases showing 
the ethical and damaging consequences of deploying opaque vision models, 
\eg~\cite{buolamwini2018gender,hansson2021self}.
Currently, however, the explainability of spatiotemporal models is under-explored~\cite{hiley2019explainable}. Some evidence suggests that these models exhibit considerable bias toward static information, \eg~\cite{vu2014predicting,he2016human,choi2019can}; therefore, an interesting question 
to answer about 
the representations in deep spatiotemporal models is: \textit{How much static and dynamic information is being captured}? While a few video interpretation methods exist, they have various limitations, \eg being primarily qualitative~\cite{feichtenhofer2020deep}, using a certain dataset that prevents evaluating the effect of the training dataset~\cite{hadji2018new} or using classification accuracy as a metric without quantifying 
a model's \textit{internal} representations~\cite{hadji2018new, sevilla2021only}.

In response, we present a quantitative paradigm for evaluating the extent that spatiotemporal models are biased toward static or dynamic information in their internal representations.
We define bias toward a certain factor (dynamic or static) as the percentage of units (\ie channels) within intermediate layers that encode that factor; see Fig.~\ref{fig:motivation} (top). Inspired by previous work~\cite{esser2020disentangling,islam2021shape}, we propose a metric to estimate the amount of static vs.\ dynamic bias based on the mutual information between sampled video pairs corresponding to these factors. We explore two common tasks 
to show the efficacy of our approach as a general tool for understanding spatiotemporal models, action recognition and video object segmentation. We focus our study on answering the following three questions: (i) What effect does the model architecture have on static and dynamic biases? (ii) How does the training dataset affect these biases? (iii) What role do units that jointly encode static and dynamic information play in relation to the architecture and dataset? 

\noindent{\bf Contributions.} Overall, we make three main contributions. (i) We introduce a general method for quantifying the static and dynamic bias contained in spatiotemporal models, including a novel sampling procedure to produce static and dynamic video pairs. (ii) We propose a technique for identifying units that jointly encode static and dynamic factors. (iii) Using the aforementioned techniques, we provide a unified study on two widely researched tasks, action recognition and video object segmentation, with a focus on the effect of architecture and training dataset on a model's static and dynamic biases; see Fig.~\ref{fig:motivation} (bottom). Among other findings, we discover in both tasks that all networks are heavily static biased, except for two-stream architectures with cross connections encouraging models to capture dynamics. Additionally, we confirm that, contrary to previous beliefs~\cite{li2018resound,bertasius2021space}, the Diving48~\cite{li2018resound} dataset is not dynamically biased and Something-Something-v2 (SSv2)~\cite{goyal2017something} is better suited to evaluate a model's ability to capture dynamics.

\section{Related work}
\noindent \textbf{Interpretability of spatiotemporal models.}
Limited work has been dedicated to the interpretability of spatiotemporal models. Several efforts predicate model interpretation on proxy tasks, \eg dynamic texture recognition~\cite{hadji2018new} or future frame selection~\cite{ghodrati2018video}. These approaches do not interpret the learned representations in the intermediate layers and in some cases require training to be performed on specific datasets~\cite{hadji2018new}. Other work focused on understanding latent representations in spatiotemporal models either mostly concerned qualitative visualization~\cite{feichtenhofer2020deep} or a specific architecture type~\cite{zhao2021interpretable}. A related task is understanding the scene representation bias of action recognition datasets~\cite{li2018resound, li2019repair}. 
However, these efforts did not focus on the effect of different architectural inductive biases on the learned intermediate representations. Our proposed interpretability technique is
the first to \emph{quantify} static and dynamic biases on \textit{intermediate} representations learned in off-the-shelf models for multiple video-based tasks. Most prior efforts focused on a single task, and studied either datasets~\cite{li2018resound} or architectures~\cite{feichtenhofer2020deep,manttari2020interpreting}. In contrast, our unified study covers six datasets and dozens of architectures on two different tasks, \ie action recognition and video object segmentation. 





\noindent \textbf{Spatiotemporal models.} Deep spatiotemporal models that learn discriminative features across space and time have proven effective for 
video understanding tasks~\cite{aafaq2019video,zhu2020comprehensive,wang2021survey}. 
Extant models 
can be broadly categorized (agnostic of the downstream task) into: two-stream approaches that separately model motion and appearance features~\cite{carreira2017quo,jain2017fusionseg,zhou2020motion,ren2021reciprocal,feichtenhofer2019slowfast}, 
3D convolutions that jointly model motion and appearance~\cite{carreira2017quo},
attention-based models with different forms of spatiotemporal data association~\cite{bertasius2021space,ren2021reciprocal}, models relying on recurrent neural networks~\cite{tokmakov2017learning} and hybrid models that combine elements of the aforementioned models~\cite{tokmakov2017learning,carreira2017quo, ren2021reciprocal}. 
Our approach to quantifying bias is not limited to the particulars of a model and is applicable to all extant and future models. We empirically demonstrate the flexibility of our approach by evaluating a diverse set of models.

\begin{figure*}[t]
    \includegraphics[width=\textwidth]{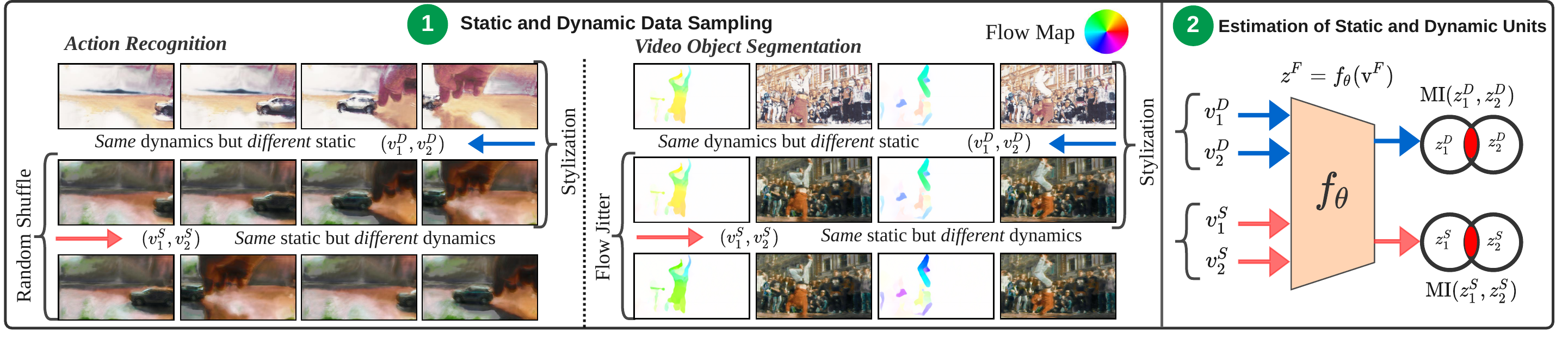}
    \vspace{-0.2cm}
    \caption{Overview of our method for analysing bias toward static or dynamic information. We measure the dynamic and static biases in deep spatiotemporal models for two tasks: action recognition and video object segmentation. \textbf{(1)} We sample video pairs that share either \textit{static}, $(v^S_1, v^S_2)$, or \textit{dynamic}, ($v^D_1,v^D_2$), information using video stylization~\cite{texler2020interactive} and frame shuffling or optical flow jitter (flow visualized in RGB format). \textbf{(2)} Given a pretrained model, $f_\theta$, we compute the mutual information (MI) between intermediate representations of video pairs, $z^F$, to assess the model's bias toward either factor on a per-layer, $l$, or per-channel (\ie unit) basis. In the appendix (see Sec.~\ref{sec:video}), we provide stylization examples in video format as well as additional static and dynamic samples.}
    \label{fig:mainmethod}
\end{figure*}

\noindent{\bf Action recognition.} 
3D convolutional networks are popular for learning 
spatiotemporal representations of videos for action recognition, \eg~\cite{taylor2010convolutional, ji20123d,tran2015learning,carreira2017quo,hara2017learning}. 
Other work has considered two-stream architectures,
where the dynamics were provided directly to one of the streams as optical flow, \eg~\cite{simonyan2014two,feichtenhofer2017spatiotemporal}.
Representative of the state of the art with convolutional networks is SlowFast~\cite{feichtenhofer2019slowfast}, which
is a two-stream 3D CNN that only takes RGB videos as input.
To encourage each stream to specialize in capturing predominately static or dynamic information, 
the temporal sampling rates of the inputs to each stream differ.
Recently, attention based approaches have proven to be suited to both static and time-series visual data, including action recognition, with variants of the transformer architecture~\cite{vaswani2017attention,bertasius2021space,fan2021multiscale,patrick2021keeping}.


\noindent{\bf Video object segmentation.} Deep video object segmentation (VOS) approaches can be categorized 
as automatic, semi-automatic and interactive~\cite{wang2021survey}.
In this work, we focus on automatic approaches that segment salient objects in videos, and the related task of motion segmentation~\cite{dave2019towards}.
We consider
two-stream models that fuse motion and appearance features. We also investigate the effect of no cross connections~\cite{jain2017fusionseg} relative to both motion-to-appearance~\cite{zhou2020motion} or bidirectional~\cite{ren2021reciprocal} cross connections.

\section{Methodology}
We introduce a novel approach to quantify the number of units (\ie channels in a given layer) encoding static and dynamic information in spatiotemporal models; for an overview, see 
Fig.~\ref{fig:mainmethod}. Our approach consists of two main steps. First, given a number of pretrained spatiotemporal models on various datasets, we sample static and dynamic pairs of videos (Sec.~\ref{sec:sampling}). Second, we use these static and dynamic pairs to estimate the number of units in the model encoding each factor based on the mutual information shared between the pairs (Sec.~\ref{sec:mmi}). 

\subsection{Sampling static and dynamic pairs}\label{sec:sampling}
\noindent \textbf{Why static and dynamic?}
We define static as `information arising from single frames' and dynamic as `information arising from the consideration of multiple frames'. The main alternative attribute to dynamics that we considered was `image motion' (\ie trackable points or regions), but `motion' is a subset of dynamic information~\cite{wildes2000qualitative,derpanis2011spacetime} (\eg stationary flashing lights have dynamics but no motion). Thus, we consider dynamics over motion because it encompasses a wider range of visual phenomena. In complement, we choose the term `static' over the possible alternative `appearance', because dynamics also can provide appearance information,~\eg the contour of an object, even if camouflaged in a single frame, can be revealed through its motion.
For our estimation technique, we produce video pairs that contain the same static information and perturbed dynamics, or vice versa, with the end goal of analyzing models trained on large-scale real-world datasets. We now detail our static and dynamic sampling techniques for both action recognition and VOS, as visualized in Fig.~\ref{fig:mainmethod} (panel 1).

\noindent \textbf{Action recognition.} The action recognition models we consider take in multiple frames (four to 32). To construct video pairs with the \textit{same} dynamics but \textit{different} static information (\ie~\textit{dynamic pairs}), we consider the same video but with two \textit{different} video styles. For video stylization, we use a recent video stylization method (with four possible styles) that perturbs static attributes like color, pixel intensity and texture~\cite{texler2020interactive}, but has less temporal artifacts (\eg flicker) than stylization methods that consider each image independently~\cite{huang2017arbitrary}. These video pairs will contain objects and scenes that have identical dynamics, but have perturbed static information. To construct pairs with the \textit{same} static information but \textit{different} dynamics (\ie~\textit{static pairs}), we take two videos of the same style, but randomly \textit{shuffle} the frames along the temporal axis; see Fig.~\ref{fig:mainmethod} (panel 1, left). In this case, the temporal correlations are altered while the static (\ie per-frame) information remains identical.

\noindent \textbf{Video object segmentation.} The VOS models considered~\cite{jain2017fusionseg,zhou2020motion,ren2021reciprocal} take a single RGB frame and an optical flow frame as input to the appearance and motion streams, resp.; 
see Fig.~\ref{fig:mainmethod} (panel 1, right). Therefore, we apply an alternative method to frame shuffling to obtain the \textit{static} pairs. For the \textit{static} pair, we use RGB images with the \textit{same} style but alter the dynamics by jittering the optical flow. 
The RGB flow representation is used with hue and saturation encoding direction and magnitude, resp., and it is those parameters that we jitter.
For the \textit{dynamic} pairs, we use the \textit{same} optical flow but a \textit{different} image style. For creating stylized images, we use the same video stylization method noted above for action recognition~\cite{texler2020interactive}, and then sample frames from the generated video. 


\subsection{Estimating static and dynamic units} \label{sec:mmi}

We seek to quantify the number of units (\ie \textit{channels}) in a layer encoding \textit{static} or \textit{dynamic} information as well as the extent to which individual units perform static, dynamic or joint encodings. Inspired by recent work that focused on single images~\cite{esser2020disentangling,islam2021shape}, we use 
a mutual information estimator to measure the information shared between video pairs. 

\noindent\textbf{Layer-wise metric.} Given a pre-trained network, $f_\theta$, and a pair of videos, $v^F_1$
and $v^F_2$, that share the semantic factor $F$ (\ie \textit{static} or \textit{dynamic}), we compute the features for an intermediate layer $l$ as $z^F_1 = f^l_{\theta}(v^F_1)$ and $z^F_2 = f^l_{\theta}(v^F_2)$ (omitting the $l$ on $z$ to simplify the notation). We use $z^F_1(i), z^F_2(i)$ to denote the $i^{\text{th}}$ unit (\ie channel) in $N^l$ dimensional features after a global average pooling layer. Our guiding intuition for this measurement is that units biased toward the \textit{static} factor will result in a higher correlation among \textit{static} pairs than the \textit{dynamic} pairs and vice versa. Under the assumption that units in the intermediate representation $z^F_1(i), z^F_2(i)$ across the dataset are jointly Gaussian, the correlation coefficient can be used as a lower bound on mutual information~\cite{kraskov2004estimating,foster2011lower}, as used in previous work~\cite{esser2020disentangling,islam2021shape}.
The number of units encoding factor $F$, $N_F$, is obtained by computing the correlation coefficient, $S_F$, over all $N^l$ channels between all video pairs $z^F_1, z^F_2$, as
\begin{equation}
\begin{split}
  N_F = \sigma(\textbf{S})\cdot N^l = \frac{\exp{(S_F)}}{\sum\limits_{k=0}^K{\exp{(S_k)}}} \cdot N^l,\\
     S_F = \sum\limits_{i=1}^{N^l} \frac{\text{Covariance}(z^F_1(i), z^F_2(i))}{
        \sqrt{\text{Variance}(z^F_1(i)) \;\text{Variance}(z^F_2(i))}},
\end{split}\label{eq:biasscores}
\end{equation}
where we multiply the Softmax, $\sigma(\cdot)$, by the number of units in that layer, $N^l$, to compute the number of units encoding the semantic factor $F$ relative to the other factors considered and $K=\{{\text{static}, \text{dynamic}, \text{identical}}\}$. In addition to \textit{static} and \textit{dynamic}, we consider a third factor in (\ref{eq:biasscores}), the \textit{identical} factor, where the video pairs have the same static and dynamic factors (\ie same video, style, frame ordering and optical flow). This baseline factor is the correlation between the model's encoding of the same videos, that gives $S_\text{Identical}=1$ for all layers. 


\noindent\textbf{Unit-wise metric.} The correlation coefficient, $S_F$, estimates the relative amount of static and dynamic information over all units in a particular layer; note the pooling done by the summation \textit{before} the Softmax in the layer-wise metric, (\ref{eq:biasscores}). However, it is also desirable to measure static and dynamic information contained in each individual channel. This measurement allows for a more fine-grained analysis of how many channels (\ie units) encode a factor $F$ above a certain threshold, as well as identify any joint or residual (\ie non-dynamic or static) units. Thus, we categorize each unit based on how much information (\ie static vs.\ dynamic) is encoded, whether any units jointly encode both factors or if there are units that do not correlate with either type of information. We measure the amount of static and dynamic information encoded in each unit  $i \in {1, \dotsc, N^l}$ as
\begin{equation}
s^i_{F} = \frac{\text{Covariance}(z^F_1(i), z^F_2(i))}{\sqrt{ \text{Variance}( z^F_1(i)) \text{Variance}( z^F_2(i))}},\label{eq:bias_scores_indv}
\end{equation} 
where each $s^i_{F}$ is the information of semantic factor $F$ in unit $i$. Given these individual correlations, we calculate the individual factors by excluding the use of a Softmax and simply threshold the correlation for each factor with a constant parameter, $\lambda$, to yield our unit-wise metrics as
\begin{equation}
\begin{split}
N_{\text{Joint}} = \sum_{i=1}^{N^l}\mathbbm{1}[ s^i_{F} > \lambda \forall F \in K] \\
N_{\text{F}} = \sum_{i=1}^{N^l}\mathbbm{1}[ s^i_{F} > \lambda \land s^i_{k} < \lambda \forall k \in K, k \neq F] \\
N_{\text{Residual}} = \sum_{i=1}^{N^l}\mathbbm{1}[ s^i_F < \lambda \forall F \in K],
\end{split}
\label{eq:ind_bias_scores_diff_b}
\end{equation}
where $K=\{\text{static}, \text{dynamic}\}$, $N_{\text{Joint}}$ indicates units jointly encoding both and $N_{\text{Residual}}$ are units not correlating with these factors under a certain threshold, $\lambda$. Note that we assign units to either joint, dynamic, static or residual and do not allow for an overlap to occur. This approach allows us to investigate the existence of units that jointly encode static and dynamic factors. In all experiments, we set $\lambda=0.5$ since it is halfway between \textit{no} and \textit{full} positive correlation. See appendix (Sec.~\ref{sec:ar_vary_thresh} and~\ref{sec:vos_arch_effect_appendix}) for results with varying $\lambda$.

\section{Experimental results}
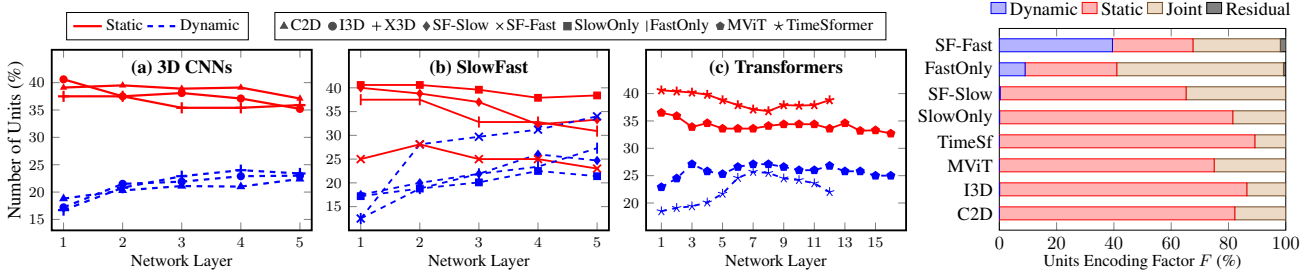
\begin{figure*} [t]
	\begin{center}
     \centering
		\resizebox{1.0\textwidth}{!}{
\begin{tikzpicture}\ref{legend_all} \ref{legend_color}
    \begin{groupplot}[group style = {group size = 3 by 1, horizontal sep = 20pt}, width = 6.0cm, height = 6.0cm]
\nextgroupplot[
      line width=1.0,
        title={\textbf{(a) 3D CNNs}},
        title style={at={(axis description cs:0.5,0.92)},anchor=north},
        xlabel={Network Layer},
        ylabel={Number of Units (\%)},
        xmin=0.8, xmax=5.2,
        ymin=13, ymax=46,
        xtick={1,2,3,4,5},
        ytick={15, 20, 25, 30, 35, 40},
        x tick label style={font=\footnotesize},
        y tick label style={font=\footnotesize},
        x label style={at={(axis description cs:0.5,0.06)},anchor=north,font=\small}, 
        y label style={at={(axis description cs:0.17,.5)},anchor=south,font=\normalsize},
        width=6.5cm,
        height=5cm,        
        ymajorgrids=false,
        xmajorgrids=false,
        major grid style={dotted,green!20!black},
    ]
    \addplot[line width=1.2pt,dashed,mark options={scale=0.8,solid},color=blue!100,mark=triangle*,]
        coordinates {(1,18.8)(2,20.3)(3,21.1)(4,21.0)(5,22.4)};
    \addplot[line width=1.2pt,mark size=1.1pt,color=red!100,mark=triangle*,]
        coordinates {(1,39.1)(2,39.5)(3,38.9)(4,39.1)(5,37.1)};
    \addplot[line width=1.2pt,dashed,mark options={scale=0.8,solid},color=blue!100,mark=*,]
        coordinates {(1,17.2)(2,21.5)(3,21.9)(4,22.9)(5,23.0)};
    \addplot[line width=1.2pt,mark options={scale=0.8,solid},color=red!100,mark=*,]
        coordinates {(1,40.6)(2,37.5)(3,38.1)(4,37.1)(5,35.2)};
    \addplot[line width=1.2pt,dashed,mark options={scale=1.5,solid},color=blue!100,mark=+,]
        coordinates {(1,16.7)(2,20.8)(3,22.9)(4,24.0)(5,23.4)};
    \addplot[line width=1.2pt,mark options={scale=1.5,solid},color=red!100,mark=+,]
        coordinates {(1,37.5)(2,37.5)(3,35.4)(4,35.4)(5,35.9)};

\nextgroupplot[
      line width=1.0,
        title={\textbf{(b) SlowFast}},
        title style={at={(axis description cs:0.5,0.92)},anchor=north,font=\normalsize},
        xlabel={Network Layer},
        xmin=0.8, xmax=5.2,
        ymin=10, ymax=48,
        xtick={1,2,3,4,5},
        ytick={15, 20, 25, 30, 35, 40},
        x tick label style={font=\footnotesize},
        y tick label style={font=\footnotesize},
        x label style={at={(axis description cs:0.5,0.06)},anchor=north,font=\small},   
        width=6.5cm,
        height=5cm,        
        ymajorgrids=false,
        xmajorgrids=false,
        major grid style={dotted,green!20!black},
        legend style={
        nodes={scale=0.87, transform shape},
        cells={anchor=west},
        legend style={at={(2,3.6)},anchor=south,row sep=0.01pt}, font =\normalsize},
        legend image post style={scale=0.9},
        legend columns=2,
        legend to name=legend_color,
    ]
    \addplot[line width=1pt,dashed,mark options={scale=0.9,solid},color=blue!100,mark=diamond*,forget plot]
        coordinates {(1,17.5)(2,20.0)(3,21.9)(4,26.0)(5,24.7)};
    \addplot[line width=1pt,mark options={scale=0.9,solid},color=red!100,mark=diamond*,forget plot]
        coordinates {(1,40.0)(2,38.8)(3,37.0)(4,32.3)(5,33.3)};
    \addplot[line width=1pt,dashed,mark options={scale=1.5,solid},color=blue!100,mark=x,forget plot]
        coordinates {(1,12.5)(2,28.1)(3,29.7)(4,31.2)(5,34.0)};
    \addplot[line width=1pt,mark options={scale=1.5,solid},color=red!100,mark=x,forget plot]
        coordinates {(1,25.0)(2,28.1)(3,25.0)(4,25.0)(5,23.0)};
    \addplot[line width=1pt,dashed,mark options={scale=0.8,solid},color=blue!100,mark=square*,forget plot]
        coordinates {(1,17.2)(2,18.8)(3,20.1)(4,22.5)(5,21.4)};
    \addplot[line width=1pt,mark options={scale=0.8,solid},color=red!100,mark=square*,forget plot]
        coordinates {(1,40.6)(2,40.6)(3,39.6)(4,37.9)(5,38.4)};
    \addplot[line width=1pt,dashed,mark options={scale=1.5,solid},color=blue!100,mark=|,forget plot]
        coordinates {(1,12.5)(2,18.8)(3,21.9)(4,23.4)(5,27.3)};
    \addplot[line width=1pt,mark options={scale=1.5,solid},color=red!100,mark=|,forget plot]
        coordinates {(1,37.5)(2,37.5)(3,32.8)(4,32.8)(5,30.9)};
    
    \addlegendimage{line width=1.2pt,color=red}\label{pgfplots:ar_stat}
    \addlegendentry[color=black]{Static}
    \addlegendimage{line width=1.2pt,dashed,color=blue}\label{pgfplots:ar_dyn}
    \addlegendentry[color=black]{Dynamic}
          
\nextgroupplot[
      line width=1.0,
        title={\textbf{(c) Transformers}},
        title style={at={(axis description cs:0.5,0.92)},anchor=north,font=\normalsize},
        xlabel={Network Layer},
        xmin=0, xmax=17,
        ymin=15, ymax=48,
        xtick={1,3,5,7,9,11,13,15},
        ytick={20, 25, 30, 35, 40},
        x tick label style={font=\footnotesize},
        y tick label style={font=\footnotesize},
        x label style={at={(axis description cs:0.5,0.06)},anchor=north,font=\small},
        width=6.5cm,
        height=5cm,         
        ymajorgrids=false,
        xmajorgrids=false,
        major grid style={dotted,green!20!black},
        legend style={
         legend style={row sep=0.1pt},
        nodes={scale=0.87, transform shape},
        legend columns=-1,
        cells={anchor=west},
        legend style={at={(10,3.6)},anchor=south,row sep=0.01pt}, font =\normalsize},
        legend to name=legend_all,
    ]
    \addplot[line width=1pt,dashed,mark options={scale=1,solid},color=blue!100,mark=pentagon*,forget plot]
        coordinates {(1,22.9)(2,24.5)(3,27.1)(4,25.8)(5,25.3)(6,26.6)(7,27.1)(8,27.1)(9,26.6)(10,26.0)(11,26.0)(12,26.8)(13,25.8)(14,25.8)(15,25.0)(16,25.0)};
    \addplot[line width=1pt,mark options={scale=1,solid},color=red!100,mark=pentagon*,forget plot]
        coordinates {(1,36.5)(2,35.9)(3,33.9)(4,34.6)(5,33.6)(6,33.6)(7,33.6)(8,34.1)(9,34.4)(10,34.4)(11,34.4)(12,33.6)(13,34.6)(14,33.2)(15,33.3)(16,32.7)};
    \addplot[line width=1pt,dashed,mark options={line width=0.5pt,scale=1.3,solid},color=blue!100,mark=star,forget plot]
        coordinates {(1,18.5)(2,19.1)(3,19.4)(4,20.1)(5,21.7)(6,24.6)(7,25.7)(8,25.5)(9,24.5)(10,24.2)(11,23.6)(12,22.0)};
    \addplot[line width=1pt,mark options={scale=1.3,solid},color=red!100,mark=star,forget plot]
        coordinates {(1,40.6)(2,40.4)(3,40.2)(4,39.8)(5,38.8)(6,37.9)(7,37.1)(8,36.8)(9,37.9)(10,37.8)(11,37.9)(12,38.8)};

    
    \addlegendimage{only marks,mark=triangle*,mark size=2.2pt,color=black!70}\label{pgfplots:ar_c1r1}
\addlegendentry[color=black]{C2D \hspace{1cm}}
    \addlegendimage{only marks,mark=*,mark size=2.1pt,color=black!70}\label{pgfplots:ar_c1r2}
\addlegendentry[color=black]{I3D  \hspace{1cm}}
    \addlegendimage{only marks,mark=+,mark size=2.5pt,color=black!90}\label{pgfplots:ar_c1r3}
\addlegendentry[color=black]{X3D  \hspace{1cm}}
    \addlegendimage{only marks,mark=diamond*,mark size=2.3pt,color=black!70}\label{pgfplots:ar_c1r4}
\addlegendentry[color=black]{SF-Slow  \hspace{1cm}}
    \addlegendimage{only marks,mark=x,mark size=2.6pt,color=black!90}\label{pgfplots:ar_c1r5}
\addlegendentry[color=black]{SF-Fast  \hspace{1cm}}
    \addlegendimage{only marks,mark=square*,mark size=2pt,color=black!70}\label{pgfplots:ar_c1r6}
\addlegendentry[color=black]{SlowOnly  \hspace{1cm}}
    \addlegendimage{only marks,mark=|,mark size=3pt,color=black!90}\label{pgfplots:ar_c1r7}
\addlegendentry[color=black]{FastOnly  \hspace{1cm}}
    \addlegendimage{only marks,mark=pentagon*,mark size=2.2pt,color=black!70}\label{pgfplots:ar_c1r8}
\addlegendentry[color=black]{MViT  \hspace{1cm}}
    \addlegendimage{only marks,mark=star,mark size=2.6pt,color=black!90}\label{pgfplots:ar_c1r9}
\addlegendentry[color=black]{TimeSformer  \hspace{1cm}}


\end{groupplot}

    
\end{tikzpicture}    
\begin{tikzpicture}
\begin{axis} [xbar stacked,
    width=7cm,
    bar width = 7pt,
    height=5.4cm,  
    xmin = 0,
    xmax = 100,
    x label style={at={(axis description cs:0.5,0.05)},anchor=north,font=\small},
    xlabel = Units Encoding Factor $F$ (\%),
    ytick=data,
    legend style={
            draw = none,
			area legend,
			at={(0.5,1.17)},
			anchor=north,
			legend columns=-1},
    symbolic y coords={C2D, I3D, MViT, TimeSf, SlowOnly, SF-Slow, FastOnly, SF-Fast},
    enlarge x limits = {value = .1},
]
 



\addplot coordinates {(0,C2D) [0] (0.1,I3D) [2] (0,MViT) [0] (0,TimeSf) [0] (0.1,SlowOnly) [2] (0.3,SF-Slow) [6] (9.0,FastOnly) [23] (39.5,SF-Fast) [101]}; 
\addplot coordinates {(82.2,C2D) [1684] (86.3,I3D) [1767] (75,MViT) [576] (89.2,TimeSf) [685] (81.4,SlowOnly) [1668] (64.9,SF-Slow) [1329] (32,FastOnly) [82] (28.1,SF-Fast) [72]}; 

\addplot coordinates {(17.8,C2D) [364] (13.6,I3D) [278] (25,MViT) [192] (10.8,TimeSf) [83] (18.5,SlowOnly) [378] (34.7,SF-Slow) [710] (58.2,FastOnly) [149] (30.5,SF-Fast) [78]}; 

\addplot coordinates {(0,C2D) [0] (0.05,I3D) [1] (0,MViT) [0] (0,TimeSf) [0] (0,SlowOnly) [0] (0.14,SF-Slow) [3] (0.8,FastOnly) [2] (2,SF-Fast) [5]}; 

\legend {Dynamic,Static,Joint,Residual};
\end{axis}
\end{tikzpicture}
} 
	\end{center}
	\vspace{-20pt}
	\caption{Layerwise and unit analyses on action recognition networks trained on Kinetics-400~\cite{carreira2017quo}. \textbf{Left:} Layerwise encoding of static and dynamic factors using the layer-wise metric, (\ref{eq:biasscores}), for: (a) single stream 3D CNNs, (b) SlowFast variants and (c) transformer variants. SF-Slow and SF-Fast denote the representation taken before the fusion layer from the slow and fast branches, resp. \textbf{Right:} Estimates of the dynamic, static, joint and residual units using the unit-wise metric, (\ref{eq:ind_bias_scores_diff_b}), on the final representation before the fully connected layer.
	}\label{fig:stagewise_ar_all_models}
	\vspace{-15pt}
\end{figure*}

We choose the two tasks of action recognition and video object segmentation to demonstrate the generality of our approach. More specifically, they differ in their semantics (\ie multi-class vs.\ binary classification), labelling (\ie video-level vs.\ pixel-level), and input types (multi-frame images vs.\ single frame optical flow). We explore three main research questions and show the corresponding results with respect to our quantitative techniques for both tasks: (i) What is the effect of the model architecture on the \textit{static} and \textit{dynamic} biases (Sec.~\ref{sec:architectures})? (ii) What effect does the training dataset have on \textit{static} and \textit{dynamic} biases (Sec.~\ref{sec:dataset_effect})? (iii) What are the characteristics of jointly encoding units in relation to model architectures and datasets? Training and implementation details can be found in appendix Sections~\ref{sec:impAR} and~\ref{sec:impVOS}. 


\subsection{What effect does model architecture have on static and dynamic biases?}
\label{sec:architectures}

\subsubsection{Action recognition}\label{sec:ar_models}
\noindent\textbf{Architectures.} As the field of action recognition has largely moved away from explicit input motion representations (\eg optical flow), we restrict our analysis to models that solely use the RGB modality. We study three types of models with respect to their static and dynamic biases: (i) single stream 3D CNNs (\ie C2D~\cite{wang2018non}, I3D~\cite{carreira2017quo} and X3D~\cite{feichtenhofer2020x3d} models), (ii) SlowFast~\cite{feichtenhofer2019slowfast} variations, where we also study the two streams when trained individually, referred to as the SlowOnly and FastOnly models and (iii) transformer-based architectures~\cite{fan2021multiscale,bertasius2021space}. All models in this subsection are trained on the Kinetics-400 dataset~\cite{carreira2017quo} and taken from the SlowFast repository~\cite{feichtenhofer2019slowfast} without any training on our part (except FastOnly, which we implement). For all models, the number of frames and sampling rate is ($8\times8$), except for the FastOnly network ($32\times2$), MViT ($16\times4$) and TimeSformer ($8\times32$). To identify the static and dynamic units of all models, we generate the Stylized ActivityNet~\cite{caba2015activitynet} validation set and use it for sampling \textit{static} and \textit{dynamic} pairs. We choose this dataset since the action distribution is similar to Kinetics-400, yet much smaller in size making it memory efficient when computing (\ref{eq:biasscores}) and (\ref{eq:ind_bias_scores_diff_b}).

\begin{figure*} [t]
	\begin{center}
     \centering 
     \begin{minipage}{0.7\textwidth}
		\resizebox{\textwidth}{!}{
     
     \begin{tikzpicture} \ref{legend_color_vos} \ref{legend_shades_vos}
     \begin{groupplot}[group style = {group size = 3 by 1, horizontal sep = 20pt}, width = 6.0cm, height = 5.0cm]
     \nextgroupplot[
                 line width=1.0,
                 title={\textbf{(a) Appearance Stream}},
                 title style={at={(axis description cs:0.5,0.93)},anchor=north,font=\normalsize},
                 xlabel={Network Layer},
                 ylabel={Number of Units (\%)},
                 xmin=0.5, xmax=5.5,
                 ymin=12, ymax=49,
                 xtick={1,2,3,4,5},
                 ytick={15,25,35,45},
                 x tick label style={font=\footnotesize},
                 y tick label style={font=\footnotesize},
                 x label style={at={(axis description cs:0.5,0.07)},anchor=north,font=\small},
                 y label style={at={(axis description cs:0.17,.5)},anchor=south,font=\normalsize},
                 width=6.5cm,
                 height=5cm,
                 ymajorgrids=false,
                 xmajorgrids=false,
                 major grid style={dotted,green!20!black},
                 legend style={
                nodes={scale=0.87, transform shape},
                cells={anchor=west},
                legend style={at={(5.2,3.7)},anchor=south,row sep=0.01pt}, font =\normalsize},
                legend image post style={scale=0.9},
                legend columns=2,
                legend to name=legend_color_vos,
             ]
            \addlegendimage{line width=1.2pt,mark=square,mark options={line width=0.9pt,scale=1.2,solid},color=blue}
            \addlegendentry[color=black]{Dynamic}
            \addlegendimage{line width=1.2pt,mark=oplus,color=red,mark options={line width=0.7pt,scale=1.3,solid}}
            \addlegendentry[color=black]{Static}
            
             \addplot[line width=1.2pt, mark options={line width=0.8pt,scale=1.1,solid}, color=blue, mark=square, style=solid]
                     coordinates {(1,14.9)
                                  (2,25.6)
                                  (3,24.1)
                                  (4,25.4)
                                  (5,28.3)};

            \addplot[line width=1.2pt,mark options={line width=0.5pt,scale=1.2,solid}, color=red, mark=oplus, style=solid]
                     coordinates {(1,40.3)
                                  (2,34.1)
                                  (3,35.2)
                                  (4,35.9)
                                  (5,33.4)};
            
            \addplot[line width=1.2pt,mark options={line width=0.8pt,scale=1.1,solid}, color=blue, mark=square, style=dashed]
                     coordinates {(1,14.9)
                                  (2,16.4)
                                  (3,16.3)
                                  (4,18.5)
                                  (5,18.2)};
            \addplot[line width=1.2pt,mark options={line width=0.5pt,scale=1.2,solid}, color=red, mark=oplus, style=dashed]
                     coordinates {(1,40.6)
                                  (2,41.1)
                                  (3,41.4)
                                  (4,40.3)
                                  (5,40.6)};
                                  
            \addplot[line width=1.2pt,mark options={line width=0.8pt,scale=1.1,solid}, color=blue, mark=square, style=loosely dotted]
                     coordinates {(1,14.9)
                                  (2,15.8)
                                  (3,15.8)
                                  (4,19.0)
                                  (5,19.3)};
                                  
             \addplot[line width=1.2pt, mark options={line width=0.5pt,scale=1.2,solid}, color=red, mark=oplus, style=loosely dotted]
                     coordinates {(1,40.6)
                                  (2,41.0)
                                  (3,41.0)
                                  (4,39.9)
                                  (5,40.0)};
            
            \nextgroupplot[ line width=1.0,
                 title={\textbf{(b) Motion Stream}},
                 title style={at={(axis description cs:0.5,0.92)},anchor=north,font=\normalsize},
                 xlabel={Network Layer},
                 ylabel={},
                 xmin=0.5, xmax=5.5,
                 ymin=12, ymax=49,
                 xtick={1,2,3,4,5},
                ytick={15,25,35,45},
                x tick label style={font=\footnotesize},
                 y tick label style={font=\footnotesize},
                 x label style={at={(axis description cs:0.5,0.07)},anchor=north,font=\small},
                 width=6.5cm,
                 height=5cm,
                 ymajorgrids=false,
                 xmajorgrids=false,
                 major grid style={dotted,green!20!black},
                 legend style={
                     legend style={row sep=0.1pt},
                    nodes={scale=0.87, transform shape},
                    legend columns=-1,
                    cells={anchor=west},
                    legend style={at={(10,3.7)},anchor=south,row sep=0.01pt}, font =\normalsize},
                    legend to name=legend_shades_vos,
                 ]
             
            \addlegendimage{line width=1.2pt,color=black, style=solid}
            \addlegendentry[color=black]{RTNet}
            \addlegendimage{line width=1.2pt,color=black, style=dashed}
            \addlegendentry[color=black]{MATNet}
            \addlegendimage{line width=1.2pt,color=black, style=loosely dotted}
            \addlegendentry[color=black]{FusionSeg}
            
            \addplot[line width=1.2pt, mark options={line width=0.8pt,scale=1.1,solid}, color=blue, style=loosely dotted, mark=square]
                     coordinates {(1,38.7)
                                  (2,39.3)
                                  (3,38.9)
                                  (4,37.1)
                                  (5,38.6)};
                                 
            \addplot[line width=1.2pt, mark options={line width=0.8pt,scale=1.1,solid}, color=blue, style=dashed, mark=square]
                     coordinates {(1,35.9)
                                  (2,36.7)
                                  (3,36.5)
                                  (4,36.3)
                                  (5,36.4)};
            \addplot[line width=1.2pt, mark options={line width=0.8pt,scale=1.1,solid}, color=blue, style=solid, mark=square]
                     coordinates {(1,34.0)
                                  (2,30.1)
                                  (3,28.9)
                                  (4,31.2)
                                  (5,32.1)};
                                  
            \addplot[line width=1.2pt, mark options={line width=0.5pt,scale=1.2,solid}, color=red, style=loosely dotted, mark=oplus]
                     coordinates {(1,21.1)
                                  (2,21.6)
                                  (3,21.2)
                                  (4,20.5)
                                  (5,20.7)};
            \addplot[smooth, line width=1.2pt, mark options={line width=0.5pt,scale=1.2,solid}, color=red, style=dashed, mark=oplus]
                     coordinates {(1,22.3)
                                  (2,22.8)
                                  (3,22.7)
                                  (4,23.2)
                                  (5,24.7)};
            
            \addplot[line width=1.2pt,  mark options={line width=0.5pt,scale=1.2,solid}, color=red, style=solid, mark=oplus]
                     coordinates {(1,23.4)
                                  (2,25.9)
                                  (3,25.8)
                                  (4,27.6)
                                  (5,28.8)};
        
        \nextgroupplot[ line width=1.0,
                 title={\textbf{(c) Fusion}},
                 title style={at={(axis description cs:0.5,0.92)},anchor=north,font=\normalsize},
                 xlabel={Network Layer},
                 ylabel={},
                 xmin=1.5, xmax=5.5,
                 ymin=12, ymax=49,
                 xtick={1,2,3,4,5},
                ytick={15,25,35,45},
                x tick label style={font=\footnotesize},
                 y tick label style={font=\footnotesize},
                 x label style={at={(axis description cs:0.5,0.07)},anchor=north,font=\small},
                 width=6.5cm,
                 height=5cm,
                 ymajorgrids=false,
                 xmajorgrids=false,
                 major grid style={dotted,green!20!black},
             ]
            \addplot[line width=1.2pt, mark options={line width=0.9pt,scale=1.1,solid}, color=blue, style=loosely dotted, mark=square]
                     coordinates {(2,33.9)
                                  (5,27.5)};
                                 
            \addplot[line width=1.2pt, mark options={line width=0.9pt,scale=1.1,solid}, color=blue, style=dashed, mark=square]
                     coordinates {(2,32.1)
                                  (3,26.5)
                                  (4,23.0)
                                  (5,23.1)};
             \addplot[line width=1.2pt, mark options={line width=0.9pt,scale=1.1,solid}, color=blue, mark=square]
                     coordinates {(2,20.7)
                                  (3,20.9)
                                  (4,21.4)
                                  (5,28.0)};
                                  
            \addplot[line width=1.2pt,  mark options={line width=0.5pt,scale=1.2,solid}, color=red, style=loosely dotted, mark=oplus]
                     coordinates {(2,25.1)
                                  (5,30.5)};
            \addplot[line width=1.2pt,   mark options={line width=0.5pt,scale=1.2,solid}, color=red, style=dashed, mark=oplus]
                     coordinates {(2,26.2)
                                  (3,30.6)
                                  (4,35.2)
                                  (5,35.5)};
            
            \addplot[line width=1.2pt,  mark options={line width=0.5pt,scale=1.2,solid}, color=red, mark=oplus]
                     coordinates {(2,32.8)
                                  (3,35.0)
                                  (4,36.7)
                                  (5,33.1)};

\end{groupplot}
\vspace{-0.5cm}
        
\end{tikzpicture}
}
\end{minipage}%
\begin{minipage}{0.3\textwidth}
\resizebox{\textwidth}{!}{
\begin{tikzpicture}
\begin{axis} [
    width=\axisdefaultwidth,
    height=4cm,
    xbar stacked,
    bar width = 10pt,
    xmin = 0,
    xmax = 100,
    xlabel = Units Encoding Factor $F$ (\%),
    ytick=data,
    legend style={
            draw = none,
			area legend,
			at={(0.5,1.3)},
			anchor=north,
			legend columns=-1},
    symbolic y coords={FusionSeg, MATNet, MATNet NoBAR, RTNet},
    enlarge x limits = {value = .1},
]
\addplot coordinates {(8.7890625,FusionSeg) (37.6953125,MATNet) (40.185546875,MATNet NoBAR) (0.0,RTNet)};
\addplot coordinates {(17.08984375,FusionSeg) (47.8515625,MATNet) (45.3857421875,MATNet NoBAR) (1.171875,RTNet)};
\addplot coordinates {(73.486328125,FusionSeg) (14.453125,MATNet) (14.4287109375,MATNet NoBAR) (98.828125,RTNet)};
\addplot coordinates {(0.634765625,FusionSeg) (0.0,MATNet) (0.0,MATNet NoBAR) (0.0,RTNet)};

\legend {Dynamic,Static,Joint,Residual};

\end{axis}
\end{tikzpicture}
}
\end{minipage}
	\end{center}
	\vspace{-20pt}
	\caption{Layer and unit-wise analysis on off-the-shelf VOS networks. \textbf{Left}: Encoding of dynamic and static factors for motion, appearance streams and fusion layers in FusionSeg~\cite{jain2017fusionseg}, MATNet~\cite{zhou2020motion} and RTNet~\cite{ren2021reciprocal} using the layer-wise metric, (\ref{eq:biasscores}). Fusion layers are mostly biased toward the static factor. \textbf{Right}: 
	Unit analysis for the three models targeting fusion layer five using the unit-wise metric, (\ref{eq:ind_bias_scores_diff_b}). MATNet has the largest number of dynamic units. MATNet NoBAR represents MATNet without the boundary-aware refinement module.}\label{fig:stagewise_vos}
	
	\vspace{-15pt}
\end{figure*}
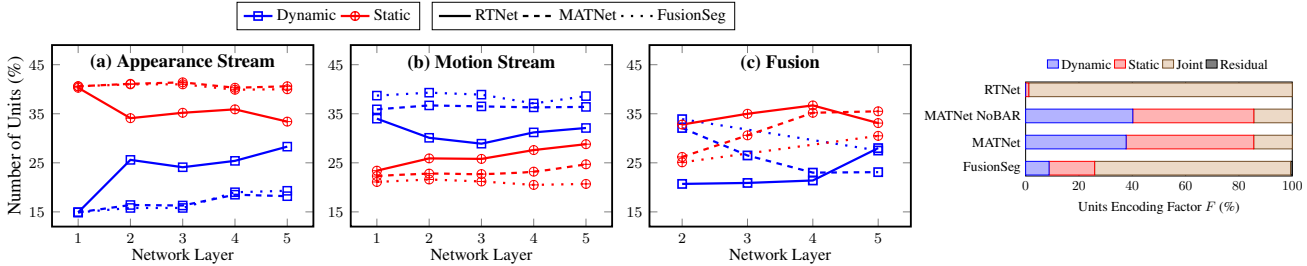

\noindent \textbf{Layer-wise analysis.} 
The static and dynamic units of multiple spatiotemporal models are quantified in Fig.~\ref{fig:stagewise_ar_all_models} (left) using our layer-wise metric, (\ref{eq:biasscores}). While the transformers are measured at every layer, the convolutional architectures are measured at five `stages', corresponding to ResNet-50-like blocks~\cite{he2016deep}. We begin our examination by comparing the last layer (\ie stage five) of each model, as this representation contains the final information before the model output. Interestingly, all single stream networks other than the \mbox{FastOnly} model are heavily biased toward \textit{static} information even though the video frames of the static pairs are \textit{randomly shuffled}. This result demonstrates the heavy bias toward static feature representations in these models. In fact, most of the 3D CNNs (\eg I3D and SlowOnly) have a similar percentage of dynamic units as the C2D network, suggesting that these models do not sufficiently capture complex dynamic representations.

We perform the static and dynamic estimation on the representations for the slow and the fast branch of the SlowFast model separately (\ie before fusion of the features). As shown in Fig.~\ref{fig:stagewise_ar_all_models} (b), this dual-stream technique for capturing dynamic information works well, as the fast branch has a significant number of dynamic units, even without the use of optical flow as input. Notably, this finding also holds for other datasets as well (see Sec.~\ref{sec:dataset_effect}). One key component of the SlowFast network is the fusion branch that aims to transfer information from the fast branch to the slow branch. This operations is performed by concatenating the slow and fast features followed by a time-strided convolution. Since the SlowOnly network is simply the SlowFast network without the fast branch, comparing the dynamic and static between the SlowOnly and SlowFast (slow) branch can reveal whether dynamic information is transferred between the pathways. The addition of the fast pathway increases the dynamic units in the slow pathway by 3.3\% as early as stage two. Additional experiments in the appendix, Sec.~\ref{sec:ar_frame}, show the robustness of our conclusion with a varying number of input frames and sampling rates. 


Looking beyond solely the final layer of the models reveal a number of interesting observations. Fig.~\ref{fig:stagewise_ar_all_models} demonstrates how all models are biased toward \textit{static} information at the earlier layers, with a tendency to encode more dynamics deeper in the network. The C2D, I3D and X3D models have only small, generally monotonic, changes in dynamic and static information at each stage. The SlowFast-Fast branch has the largest change in terms of the dynamic units, again showing the ability of the two-stream architecture to capture dynamic information. Conversely, the per-layer characteristics of static and dynamic encoding is different in both transformer-based architectures. They encode an increasing amount of dynamic information up until about halfway through the model, at which point the pattern tapers off and even reverses slightly. 

\noindent \textbf{Unit-wise analysis.} 
We now examine individual units using our unit-wise metric, (\ref{eq:ind_bias_scores_diff_b}),
with $\lambda=0.5$ and report the results for the final representation before the fully connected layer in Fig.~\ref{fig:stagewise_ar_all_models} (right). Interestingly, all single stream models, other than FastOnly, contain mainly \textit{static} and \textit{joint} units. There appears to be no difference between single-stream transformers and CNNs in the emergence of dynamic or residual units. In contrast, the FastOnly model and SlowFast-Fast branch produce a significant number of \textit{dynamic} units. Another finding consistent with the results from Fig.~\ref{fig:stagewise_ar_all_models} (right), is revealed when comparing the \mbox{FastOnly} model and SlowFast-Fast branch: The Fast model extracts more dynamic information \textit{when trained jointly with the Slow branch}. These findings all together demonstrate the efficacy of two-stream architectures with varying capacity and frame rates. 
In the appendix (see Sec.~\ref{sec:ar_vary_thresh}), we verify 
that this pattern of results remain consistent while varying the threshold, $\lambda$, and provide results at multiple layers.


\subsubsection{Video object segmentation} 
\noindent\textbf{Architectures.} We study the dynamic and static biases of two-stream fusion VOS models that take two-frame optical flow and an RGB image as input, with different types of cross connections: (i) FusionSeg~\cite{jain2017fusionseg} with no cross connections, (ii) MATNet~\cite{zhou2020motion} with motion-to-appearance cross connections and (iii) RTNet~\cite{ren2021reciprocal} with bidirectional cross connections. For a fair comparison with the two other models that fuse motion and appearance in the intermediate representations, we use a modified version of FusionSeg~\cite{jain2017fusionseg} trained on DAVIS16~\cite{Perazzi2016} in our analysis. Our modified model follows an encoder-decoder approach~\cite{chen2018encoder}, resulting in two fusion layers as detailed in the appendix (see Sec.~\ref{sec:impVOS}). Our model achieves similar performance to the original on DAVIS16 (70.8\% vs.\ 70.7\% mIoU). For both MATNet~\cite{zhou2020motion} and RTNet~\cite{ren2021reciprocal}, we use the models provided by the authors without further fine-tuning. We provide an analysis on MATNet trained only on DAVIS16 (\ie without additional YouTube-VOS data) in the appendix (see Sec.~\ref{sec:vos_arch_effect_appendix}). We use a stylized version of DAVIS16 in our analysis to evaluate the static and dynamic biases for the previous models, with stylization according to Sec.~\ref{sec:sampling}. In the case of both motion and appearance streams, we analyse features after cross connections, if present. In the case of fusion layers, the features extracted after the spatiotemporal attention fusion in RTNet, and the features after scale sensitive attention in MATNet are used. In FusionSeg, the features after the convolutional layers fusing motion and appearance from the second and fifth ResNet stages are used.

\noindent \textbf{Layer-wise analysis.} Figure~\ref{fig:stagewise_vos} (left), shows the layerwise analysis for the motion and appearance streams as well as the fusion layers according to our layer-wise metric, (\ref{eq:biasscores}).
Similar to our finding with the action recognition models in Sec.~\ref{sec:ar_models}, the majority of the video object segmentation models are biased toward the \textit{static} factor in the fusion layers (\ie fusion layers three, four and five). We observe an increase in the dynamic bias in the appearance stream as we go deeper in the network, especially for RTNet. In contrast, the bias in the motion streams of both FusionSeg and MATNet are somewhat consistent throughout layers. Interestingly, in RTNet, the \textit{static} bias increases as the representation goes deeper in the network. This result likely stems from the bidirectional cross-connections in RTNet.

\noindent \textbf{Unit-wise analysis.} 
The individual unit analysis for these models
obtained using our unit-wise metric, (\ref{eq:ind_bias_scores_diff_b}), with $\lambda=0.5$
is shown in Fig.~\ref{fig:stagewise_vos} (right) for fusion layer five. MATNet has a nontrivial increase of dynamics biased units compared to the other models. In contrast, RTNet and FusionSeg show a greater number of jointly encoding units, coming at the expense of units biased toward the static and dynamic factors. This pattern suggests that cross connections, as present in MATNet, can lead to an increase in the specialized units that encode the static and dynamic factors in the late fusion layers. We also show MATNet trained without its boundary-aware refinement module and boundary loss, as ``MATNet NoBAR'', confirming the source behind such an increase are the motion-to-appearance cross connections. 

As with action recognition, experiments in the appendix (see Sec.~\ref{sec:vos_arch_effect_appendix}) demonstrates that our observations are robust with respect to different fusion layers, variations of the threshold, $\lambda$, and training dataset variations (\ie without YouTube-VOS). In the appendix (see Sec.~\ref{sec:vos_arch_effect_appendix}), we also demonstrate that motion-to-appearance cross connections relate to the performance for a task requiring dynamic information (\ie the segmentation of camouflaged moving objects (MoCA)~\cite{lamdouar2020betrayed}). 






\subsubsection{Summary and shared insights}
We have shown in both action recognition and video segmentation that the majority of the examined state-of-the-art models are biased toward encoding static information. We also demonstrated the efficacy of two-stream models with motion-to-appearance~\cite{zhou2020motion} (fast-to-slow~\cite{feichtenhofer2019slowfast}) cross connections to enable greater encoding of dynamic information. Finally, we documented that the final layers of dynamic biased models are capable of producing a significant amount of specialized dynamic units compared to the joint units produced by static biased models.

\subsection{How does the training dataset affect static and dynamic biases?}\label{sec:dataset_effect}
\begin{figure} [t]
 \def\arraystretch{1.35}
 \setlength\tabcolsep{2.4pt}
\centering
\begin{minipage}{0.57\linewidth}
\resizebox{1.0\textwidth}{!}{
	\begin{tabu}{c cccc}
	\tabucline[1pt]{-}
	 \multirow{2}{*}{Dataset}&  \multicolumn{2}{c}{\textbf{SlowOnly}} & \multicolumn{2}{c}{\textbf{FastOnly}} \\
	 \cline{2-3} \cline{4-5}
	  & Dyn.(\%) & Stat.(\%)& Dyn.(\%) & Stat.(\%)\\
	   \tabucline[1pt]{-}
	   Kinetics& 21.4 & 38.4 & 27.3 & 30.9 \\
	   Diving48& 23.1 & 34.0 & 23.8 & 27.3\\
	   SSv2& 28.2 & 30.7 & 31.6 & 21.9 \\
	\tabucline[1pt]{-}
	\end{tabu}}
		\end{minipage} \hfill   
	\begin{minipage}{0.4\linewidth}
	\resizebox{1.0\textwidth}{!}{
	\begin{tikzpicture} \ref{target_legend}
    \begin{axis}[
       line width=1.0,
        title={Performance on Shuffled Frames},
        title style={at={(axis description cs:0.5,0.95)},anchor=north,font=\normalsize},
        xlabel={Dataset},
        ylabel={Relative Performance Drop (\%)},
        ymin=-65, ymax=12,
        ytick={0,-10,-20,-30,-40,-50,-60},
        symbolic x coords={SSv2, Diving48, Kinetics},
        xtick=data,
        x tick label style={font=\footnotesize},
        y tick label style={font=\footnotesize},
        x label style={at={(axis description cs:0.5,0.03)},anchor=north,font=\small},
        y label style={at={(axis description cs:0.12,.5)},anchor=south,font=\small},
        width=6.7cm,
        height=5.5cm,        
        ymajorgrids=false,
        xmajorgrids=false,
        major grid style={dotted,green!20!black},
        legend style={
         nodes={scale=0.9, transform shape},
         cells={anchor=west},
         legend style={at={(3.8,0.25)},anchor=south}, font =\footnotesize},
         legend entries={[black]SlowOnly,[black]FastOnly,[black]Baseline},
        legend to name=target_legend,
    ]
    
    \addplot[only marks,mark size=3.3pt,color=orange,mark=*,]
    coordinates {(SSv2,-59.8) (Diving48,-2.0) (Kinetics,-0.5)};

        
    \addplot[only marks,mark size=3.3pt,color=blue,mark=triangle*,]
        coordinates {(SSv2,-33.6) (Diving48,-18.0) (Kinetics,-1.2)};

    \addplot[line width=1.3pt,black,dotted,sharp plot,update limits=false] 
	    coordinates {([normalized]-10,0)([normalized]10,0)};
    \end{axis}
\end{tikzpicture}}
\end{minipage} 
\vspace{-5pt}
\caption{Analyses of biases of action recognition datasets. \textbf{Left:} \textit{Dynamic} and \textit{static} dimensions using the layer-wise metric, (\ref{eq:biasscores}), for networks trained on Kinetics-400~\cite{carreira2017quo}, Diving48~\cite{li2018resound} and SSv2~\cite{goyal2017something}. \textbf{Right:} Relative percentage drop in Top 1 Accuracy (\%) for the SlowOnly and FastOnly models trained with shuffled frames with respect to the baseline (\ie standard training). SSv2 drops more in performance than Diving48 or Kinetics-400.}\label{fig:dataset_compare} \vspace{-0.4cm}
\end{figure}
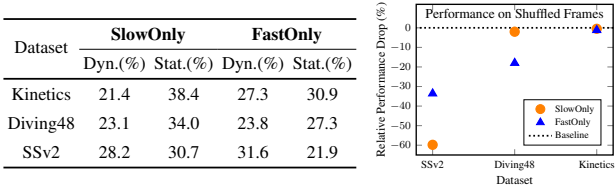

\subsubsection{Action recognition}\label{sec:ar_dataset}
\noindent \textbf{Datasets.} With the knowledge that action recognition models often use static context biases in the data to make predictions (\eg \cite{derpanis2012action,choi2019can}), we 
consider datasets in the following evaluations which were designed with the goal of benchmarking a model's ability to capture dynamic information. Two popular datasets of this type are Something-Something-v2~\cite{goyal2017something} (SSv2) and Diving48~\cite{li2018resound}. SSv2 is a fine-grained ego-centric dataset with 174 classes and over 30,000 unique objects. Notably, different actions in SSv2 include similar appearance but different motions, \eg the classes `moving something from right-to-left' and `moving something from left-to-right'. Diving48~\cite{li2018resound} was created to be ``a dataset with no significant biases toward static or short-term motion representations, so that the capability of models to capture long-term dynamics information could be evaluated''~\cite{d48_web}. All actions are a particular type of dive and differ by only a single rotation or flip. We compare Kinetics-400, Diving48 and SSv2 to determine the extent that each dataset requires dynamics for action recognition.

\noindent \textbf{Dataset bias.} We use the layerwise metric,  (\ref{eq:biasscores}), to estimate the static and dynamic units captured in the last layer of two models trained on the three datasets, as shown in the table of Fig.~\ref{fig:dataset_compare} (left). We generate Stylized SSv2 and Stylized Diving48 to produce the static and dynamic estimates (and continue using Stylized ActivityNet for Kinetics-400 trained models). We measure the last layer, as the final prediction is made directly from it and thus is most representative of what information the model uses for the final prediction. The SlowOnly and FastOnly architectures follow a similar pattern to that found in Sec.~\ref{sec:architectures}, with the FastOnly consistently capturing more dynamic information.
Surprisingly, models trained on Diving48 capture a similar amount of dynamics compared to Kinetics.
These results may seem curious at first, as it seems unlikely that models could perform well on Diving48 without dynamic information.

To further understand and confirm this result, we conduct a simple experiment, where the model only has static information to learn from. As discussed in Sec.~\ref{sec:sampling}, frame-shuffled videos will have the same static information as a non-shuffled input, but the temporal correlations, and hence dynamic information, will be corrupted. This manipulation forces the model to focus on static information for classification. We compare the top-1 validation accuracy of models trained and validated on shuffled frames to that of models with standard training. Fig.~\ref{fig:dataset_compare} (right) shows the results of the SlowOnly and FastOnly networks on Diving48, SSv2 and Kinetics-400, in terms of the relative performance on shuffled frames compared to unshuffled. For a fair comparison, we initialize all models from Kinetics-400. Both models show strong relative performance when trained to classify shuffled videos for Diving48 and Kinetics-400; however, for SSv2 the classification performance is decreased to a greater extent when trained on shuffled frames. These results show that SSv2 is a better alternative for benchmarking temporally capable networks.

\begin{figure}[t]
    \centering
    \resizebox{0.49\textwidth}{!}{
\begin{tikzpicture}
\begin{axis} [
    width=\axisdefaultwidth,
    height=3.2cm,
    xbar stacked,
    bar width = 10pt,
    xmin = 0,
    xmax = 100,
    xlabel = Units Encoding Factor $F$ (\%),
    ytick=data,
    legend style={
            draw = none,
			area legend,
			at={(0.5,1.4)},
			anchor=north,
			legend columns=-1},
    symbolic y coords={Diving48, Kinetics, SSv2},
    enlarge x limits = {value = 0.1},
]


\addplot coordinates { (0,Diving48) (0.09765625,Kinetics) (8.447265625,SSv2)};
\addplot coordinates { (89.79492188,Diving48) (81.4453125,Kinetics) (15.28320313,SSv2)};
\addplot coordinates { (0,Diving48) (18.45703125,Kinetics) (75.83007813,SSv2)};
\addplot coordinates { (10.20507813,Diving48) (0,Kinetics) (0.439453125,SSv2)};

\legend {Dynamic, Static, Joint, Residual};

\end{axis}
\end{tikzpicture}

\begin{tikzpicture}
\begin{axis} [
    width=\axisdefaultwidth,
    height=3.2cm,
    xbar stacked,
    bar width = 10pt,
    xmin = 0,
    xmax = 100,
    xlabel = Units Encoding Factor $F$ (\%),
    ytick=data,
    legend style={
            draw = none,
			area legend,
			at={(0.5,1.4)},
			anchor=north,
			legend columns=-1},
    symbolic y coords={Diving48, Kinetics, SSv2},
    enlarge x limits = {value = .1},
]


\addplot  coordinates {(10.5,Diving48) (8.9,Kinetics)  (78.1,SSv2)};
\addplot  coordinates {(14.1,Diving48)  (32.0,Kinetics) (1.5,SSv2)};
\addplot  coordinates {(0,Diving48) (58.2,Kinetics) (16.8,SSv2)};
\addplot  coordinates {(75.4,Diving48) (0.8,Kinetics) (3.5,SSv2)};

\legend {Dynamic, Static, Joint, Residual};

\end{axis}
\end{tikzpicture}
}
\vspace{-15pt}
\caption{Estimating the dynamic, static, joint and residual units using the unit-wise metric, (\ref{eq:ind_bias_scores_diff_b}), for the SlowOnly (\textbf{left}) and FastOnly (\textbf{right}) models on Kinetics-400~\cite{carreira2017quo}, Diving48~\cite{li2018resound} and SSv2~\cite{goyal2017something}. Dynamic units arise from dynamic-biased models (\eg FastOnly) and residual units from training on Diving48.}
\label{fig:ar_dataset_indiv} \vspace{-0.4cm}
\end{figure}
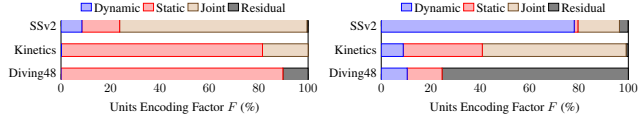

\noindent \textbf{Individual units analysis.} 
Figure~\ref{fig:ar_dataset_indiv} shows the individual units (from the last layer) for two models (one static biased, SlowOnly, and one dynamic biased, FastOnly) on Kinetics-400, Diving48 and SSv2. The SlowOnly model trained on Kinetics-400 contains only static and joint units. However, when trained on Diving48 or SSv2, both residual and dynamic units emerge, demonstrating the impact of the training dataset on producing specialized units. This finding is consistent across all static biased architectures; see appendix Sec.~\ref{sec:ar_append_dataset}. Unlike the SlowOnly model, the FastOnly model contains many dynamic units trained on any dataset, showing the efficacy of the architecture for producing specialized dynamic units. Interestingly, each dataset is unique in the type of units that emerge. Diving48 produces residual units, suggesting there are other factors at play beyond dynamic and static information. On the other hand, SSv2 produces the most dynamic units for both models. Sec.~\ref{sec:ar_append_dataset} in the appendix shows this observation is consistent with other models.

\vspace{-0.3cm}
\subsubsection{Video object segmentation} 

\begin{table}
\resizebox{0.45\textwidth}{!}{
\begin{tabu}{c cccc}
\tabucline[1pt]{-}
 \multirow{2}{*}{Dataset}&  \multicolumn{2}{c}{\textbf{Fusion Layer 5}} & \multicolumn{2}{c}{\textbf{Fusion Layer 2}}    \\
\cline{2-5}
  & Dyn.(\%) & Stat.(\%)& Dyn.(\%) & Stat.(\%)\\
\tabucline[1pt]{-}
DAVIS & 27.8 & 30.1 & 34.0 & 25.9 \\
ImageNetVID & 26.4 & 33.1 & 33.0 & 24.6\\
TAO-VOS & 26.4 & 25.8  & 33.7 & 23.2 \\
\tabucline[1pt]{-}
\end{tabu}}
\vspace{-8pt}
\caption{Biases of video object segmentation datasets using the layer-wise metric, (\ref{eq:biasscores}), for FusionSeg's fusion layers five and two, trained on DAVIS16~\cite{Perazzi2016}, ImageNetVID~\cite{jain2017fusionseg} and TAO-VOS~\cite{Voigtlaender21WACV}.}
\label{fig:vos_dataset}
\vspace{-20pt}
\end{table}

\noindent\textbf{Datasets.} We study the impact of the following three VOS datasets on a model's static and dynamic biases:
DAVIS16~\cite{Perazzi2016}, Weakly Labelled ImageNet VID~\cite{jain2017fusionseg} and TAO-VOS~\cite{Voigtlaender21WACV}.
DAVIS16~\cite{Perazzi2016} is the most widely used benchmark for automatic VOS, with 50 short-temporal extent sequences of two to four seconds and 3455 manually annotated frames. ImageNet VID~\cite{jain2017fusionseg} contains 3251 weakly labelled videos and was used in previous work to pretrain a model's motion stream~\cite{jain2017fusionseg}. Here, we use it as a general training dataset, \ie beyond just for motion streams, to assess its impact. Finally, TAO-VOS~\cite{Voigtlaender21WACV} contains 626 relatively long videos (36 seconds on average) that are annotated in a hybrid fashion between manually and weakly labelled frames, resulting in 74,187 frames.
We convert the annotations to exclude instances and instead consider foreground/background annotations only.

\noindent\textbf{Dataset bias.} We train our modified version of FusionSeg with early (layer two) and late (layer five) fusion layers on our three datasets. We compute the static and dynamic biases for the training datasets using the layer-wise metric,
(\ref{eq:biasscores}),
and report the results in Table~\ref{fig:vos_dataset}. The model trained on TAO-VOS has the least amount of static bias out of all three datasets. However, it appears that the datasets do not differ significantly in their dynamic bias. These results are further explored, by analyzing the specialized dynamic and jointly encoding units, as discussed in the next section. 

\begin{figure}
    \centering
\begin{minipage}{0.35\textwidth}
\resizebox{\textwidth}{!}{
\begin{tikzpicture}
\begin{axis} [
    width=\axisdefaultwidth,
    height=3.2cm,
    xbar stacked,
    bar width = 10pt,
    xmin = 0,
    xmax = 100,
    title = \textbf{Fusion Layer 5},
    ytick=data,
    legend style={
            draw = none,
			area legend,
			at={(0.5,1.4)},
			anchor=north,
			legend columns=-1},
    symbolic y coords={TAO-VOS, ImageNetVID, DAVIS},
    enlarge x limits = {value = .1},
] 
\addplot coordinates { (8.7890625,DAVIS) (0.9765625,ImageNetVID)  (49.70703125,TAO-VOS)};
\addplot coordinates { ( 17.08984375,DAVIS) ( 20.654296875,ImageNetVID)  (18.5546875,TAO-VOS)};
\addplot coordinates { ( 73.486328125,DAVIS) (78.173828125,ImageNetVID)  (22.021484375,TAO-VOS)};
\addplot coordinates { (0.6,DAVIS) (0.1953125,ImageNetVID)  (9.716796875,TAO-VOS)};
\end{axis}
\end{tikzpicture}}

\resizebox{\textwidth}{!}{
\begin{tikzpicture}
\begin{axis} [
    width=\axisdefaultwidth,
    height=3.2cm,
    xbar stacked,
    bar width = 10pt,
    xmin = 0,
    xmax = 100,
    title = \textbf{Fusion Layer 2},
    xlabel = Units Encoding Factor $F$ (\%),
    ytick=data,
    legend style={
            draw = none,
			area legend,
			at={(0.5,-0.7)},
			anchor=north,
			legend columns=-1},
    symbolic y coords={TAO-VOS, ImageNetVID, DAVIS},
    enlarge x limits = {value = .1},
] 
\addplot coordinates { (32.421875,DAVIS) (60.15625,ImageNetVID)  (70.703125,TAO-VOS)};
\addplot coordinates { (12.109375,DAVIS) (8.203125,ImageNetVID)  (3.515625,TAO-VOS)};
\addplot coordinates { (54.296875,DAVIS) (30.46875,ImageNetVID)  (23.828125,TAO-VOS)};
\addplot coordinates { (1.171875,DAVIS) (1.171875,ImageNetVID)  (1.953125,TAO-VOS)};
 
\legend {Dynamic, Static, Joint, Residual};
\end{axis}
\end{tikzpicture}}
\end{minipage}%
\begin{minipage}{0.13\textwidth}
\resizebox{\textwidth}{!}{
\includegraphics[width=\textwidth]{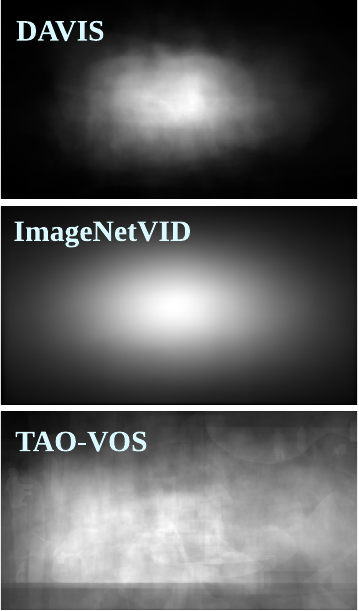}}
\end{minipage}
\vspace{-10pt}
\caption{Analyses of biases of VOS datasets. \textbf{Left}: Estimating the dynamic, static, joint and residual units using the unit-wise metric, (\ref{eq:ind_bias_scores_diff_b}), for FusionSeg's fusion layers five and two trained on DAVIS16~\cite{Perazzi2016}, ImageNetVID~\cite{jain2017fusionseg} and TAO-VOS~\cite{Voigtlaender21WACV}. \textbf{Right}: Center bias plots for the three datasets. The results show the emergence of more dynamic units for both fusion layers when trained on the least center biased dataset (\ie TAO-VOS).}
\label{fig:vos_dataset_indiv}
\vspace{-15pt}
\end{figure}
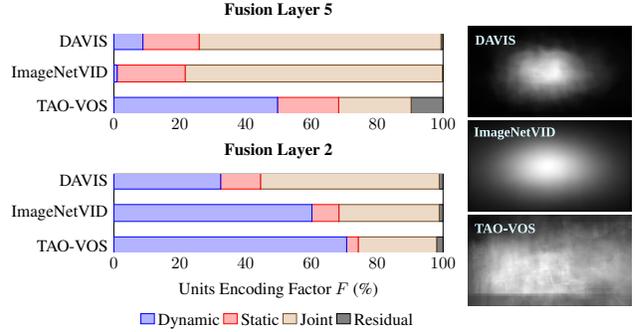

\noindent\textbf{Individual units analysis.} We analyse the datasets in terms of the individual unit analysis using 
the unit-wise metric, (\ref{eq:ind_bias_scores_diff_b}), with $\lambda=0.5$. 
It is seen in Fig.~\ref{fig:vos_dataset_indiv} (left) that models trained on TAO-VOS produce the highest number of specialized dynamic biased units, unlike DAVIS16 and ImageNet VID that show more joint units. To explore this matter further, we evaluate the center bias for the three datasets by calculating the average (normalized to 0-1) number of groundtruth segmentation masks for each pixel over the entire dataset, with results shown in Fig.~\ref{fig:vos_dataset_indiv} (right). It is seen that for both layers, the percentage of specialized dynamic units is greatest for the dataset that has least center bias, \ie TAO-VOS, as its center bias map is far more diffuse than the others.
These observations have implications for how the datasets can be used best for different tasks. For example, more general motion segmentation without concern for centering, might be better served by training with a dynamic biased dataset (\eg TAO-VOS) unlike static biased datasets (\eg DAVIS16 and ImageNet VID).

\vspace{-0.3cm}
\subsubsection{Summary and shared insights}
We have shown the effect of training datasets on both tasks. Our results raise questions about some of the widely adopted datasets in action recognition. In particular, Diving48 is claimed to be a good benchmark for learning dynamics~\cite{li2018resound}. Instead, our results suggest that SSv2 is better suited for evaluating a model's ability to capture dynamics. In video object segmentation, we found training on TAO-VOS yields the largest number of specialized dynamic units. Thus, it may be a better training dataset for tasks that rely on capturing dynamics (\eg motion segmentation).

\section{Conclusion}
This paper has advanced the understandability of learned spatiotemporal models for video understanding, especially action recognition and video object segmentation. We have introduced a general method for analyzing the extent that various architectures capitalize on static vs.\ dynamic information. We also showed how our method can be applied to investigate the static vs.\ dynamic biases in datasets. Future work can apply our method to additional video understanding tasks (\eg action prediction) as well as use insights gained on particular models and datasets to improve their performance and applicability (\eg reduce identified biases for better generalization to new data).

{\small
\noindent{\bf Acknowledgements.} We gratefully acknowledge financial support from the Canadian NSERC Discovery Grants and Vector Institute Post-graduate Affiliation award. K.G.D.\ and R.P.W.\ contributed to this work in their personal capacity as Associate Professors at York University.
}

\section{Appendix}
\subsection{Introduction}

Our appendix is organized in five major sections. Section~\ref{sec:video} documents an associated supplementary video. Sections~\ref{sec:actions} and~\ref{sec:vos} provide details regarding our action recognition and video object segmentation experiments, respectively. Each of these sections is partitioned into an initial subsection that presents implementation details, followed by a series of subsections providing additional empirical results and analyses. Finally, Section~\ref{sec:assets} documents all assets employed in our work. All references to equations refer to equations defined in the main paper.

\subsection{Demo video}\label{sec:video}

We include an accompanying demo video which can also be found on our project page\footnote{\url{https://yorkucvil.github.io/Static-Dynamic-Interpretability/}}. In this video, we show examples of the static and dynamic pairs for both action recognition and video object segmentation (VOS). The video is in MP4 format and approximately three minutes long. Layouts for each sampling pair are described in detail followed by the example video samples. The codec used for the realization of the provided video is H.264 (x264).

\subsection{Action recognition}\label{sec:actions}
In this section, we provide details for action recognition. We begin by presenting implementation details for all models evaluated in the main submission. Subsequently, we provide a supplementary series of experiments where we consider various frame rates as input to the SlowFast network~\cite{feichtenhofer2019slowfast}, variation of the threshold, $\lambda$, in the unit-wise metric, (\ref{eq:ind_bias_scores_diff_b}), and the effect of the training dataset.

\subsubsection{Implementation details}\label{sec:impAR}

The main repository used for our action recognition experiments is the SlowFast~\cite{feichtenhofer2019slowfast} repository\footnote{\url{https://github.com/facebookresearch/SlowFast}}. This repository contains dozens of pre-trained action recognition architectures trained on multiple datasets. The only model taken from a different repository is the TimeSformer~\cite{bertasius2021space}, which has its own codebase\footnote{\url{https://github.com/facebookresearch/TimeSformer}} built upon the SlowFast repository. For all models, we use the standard configuration files provided by the repository except for the following. 

The FastOnly model is implemented by us based on the SlowFast architecture found in the SlowFast repository. For a fair comparison with the SlowFast model, the FastOnly model is implemented using the same frame and sampling rate as the SlowFast-Fast branch (32 total frames sampled every two frames). 

All models trained on Kinetics-400~\cite{carreira2017quo} and Something-Something-v2~\cite{goyal2017something} (SSv2) are taken directly from the SlowFast repository, except for the FastOnly model. The FastOnly model is trained on Kinetics-400 for 40 epochs with SGD, a weight decay of 1e-4, a batch size of 32 and a base learning rate of 0.03 that decreases by a factor of 10 at epochs 15, 30 and 35. On SSv2, the FastOnly model is trained for 25 epochs with SGD, weight decay of 1e-4, a batch size of 32 and a base learning rate that is decreased by a factor of 10 at epochs 10 and 20. 

All models trained on Diving48~\cite{li2018resound} are trained by us. The FastOnly model is trained on Diving48 for 100 epochs with SGD, weight decay of 1e-4, a batch size of 32 and a base learning rate of 0.0375 that decreases by a factor of 10 at epochs 40, 60 and 80. The SlowOnly model is trained on Diving48 for 100 epochs with SGD, weight decay of 1e-4, a batch size of 32 and a base learning rate of 0.00375 that decreases by a factor of 10 at epochs 40, 60 and 80. All models trained with temporal frame shuffling (see Sec.~\ref{sec:ar_dataset} of the main paper) are trained with the same hyperparameters as their unshuffled counterparts.

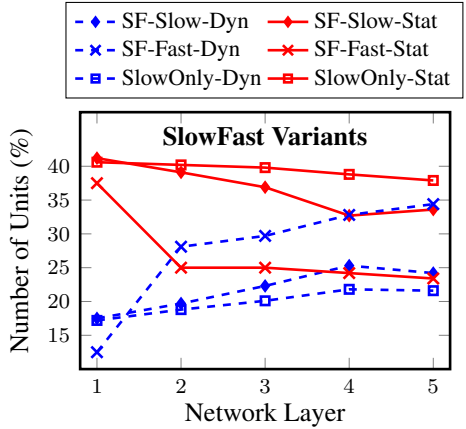
\begin{figure}[t]
    \centering
    \begin{tikzpicture} \ref{legend_4x16}
    \begin{axis}[
        line width=1.0,
        title={\textbf{SlowFast Variants}},
        title style={at={(axis description cs:0.5,0.92)},anchor=north,font=\normalsize},
        xlabel={Network Layer},
        x label style={font=\small},
        ylabel={Number of Units (\%)},
        xmin=0.8, xmax=5.2,
        ymin=10, ymax=48,
        xtick={1,2,3,4,5},
        ytick={15, 20, 25, 30, 35, 40},
        x tick label style={font=\footnotesize},
        y tick label style={font=\footnotesize},
        x label style={at={(axis description cs:0.5,0.06)},anchor=north,font=\normalsize},
        y label style={at={(axis description cs:0.15,.5)},anchor=south,font=\normalsize},
        width=6.5cm,
        height=5cm,        
        ymajorgrids=false,
        xmajorgrids=false,
        major grid style={dotted,green!20!black},
        legend style={
        nodes={scale=0.87, transform shape},
        cells={anchor=west},
        legend style={at={(2.45,3.6)},anchor=south,row sep=0.01pt}, font =\normalsize},
        legend image post style={scale=0.9},
        legend columns=2,
        legend to name=legend_4x16,
    ]

    \addplot[line width=1pt,dashed,mark options={scale=0.9,solid},color=blue!100,mark=diamond*]
        coordinates {(1,17.5)(2,19.7)(3,22.3)(4,25.3)(5,24.2)};
    \addlegendentry[black]{SF-Slow-Dyn}
    \addplot[line width=1pt,mark options={scale=0.9,solid},color=red!100,mark=diamond*]
        coordinates {(1,41.2)(2,39.1)(3,36.9)(4,32.7)(5,33.6)};
    \addlegendentry[black]{SF-Slow-Stat}
    \addplot[line width=1pt,dashed,mark options={scale=1.5,solid},color=blue!100,mark=x]
        coordinates {(1,12.5)(2,28.1)(3,29.7)(4,32.8)(5,34.4)};
    \addlegendentry[black]{SF-Fast-Dyn}
    \addplot[line width=1pt,mark options={scale=1.5,solid},color=red!100,mark=x]
        coordinates {(1,37.5)(2,25.0)(3,25.0)(4,24.2)(5,23.4)};
    \addlegendentry[black]{SF-Fast-Stat}


    \addplot[line width=1pt,dashed,mark options={scale=0.8,solid},color=blue!100,mark=square]
        coordinates {(1,17.2)(2,18.8)(3,20.1)(4,21.8)(5,21.6)};
    \addlegendentry[black]{SlowOnly-Dyn}
    \addplot[line width=1pt,mark options={scale=0.8,solid},color=red!100,mark=square]
        coordinates {(1,40.6)(2,40.2)(3,39.8)(4,38.8)(5,37.9)};
    \addlegendentry[black]{SlowOnly-Stat}
    
    \end{axis}
    \end{tikzpicture}
    \caption{Static and dynamic bias analysis on SlowFast variants with alternative sampling rates trained on Kinetics-400~\cite{carreira2017quo} using the layer-wise metric, (\ref{eq:ind_bias_scores_diff_b}). All models are trained with four frames sampled every 16 frames (\ie $4 \times 16)$. SF-Slow and SF-Fast denote the representation taken before the fusion layer from the Slow and Fast branches, respectively.}
    \label{fig:4x16_sf}
\end{figure}

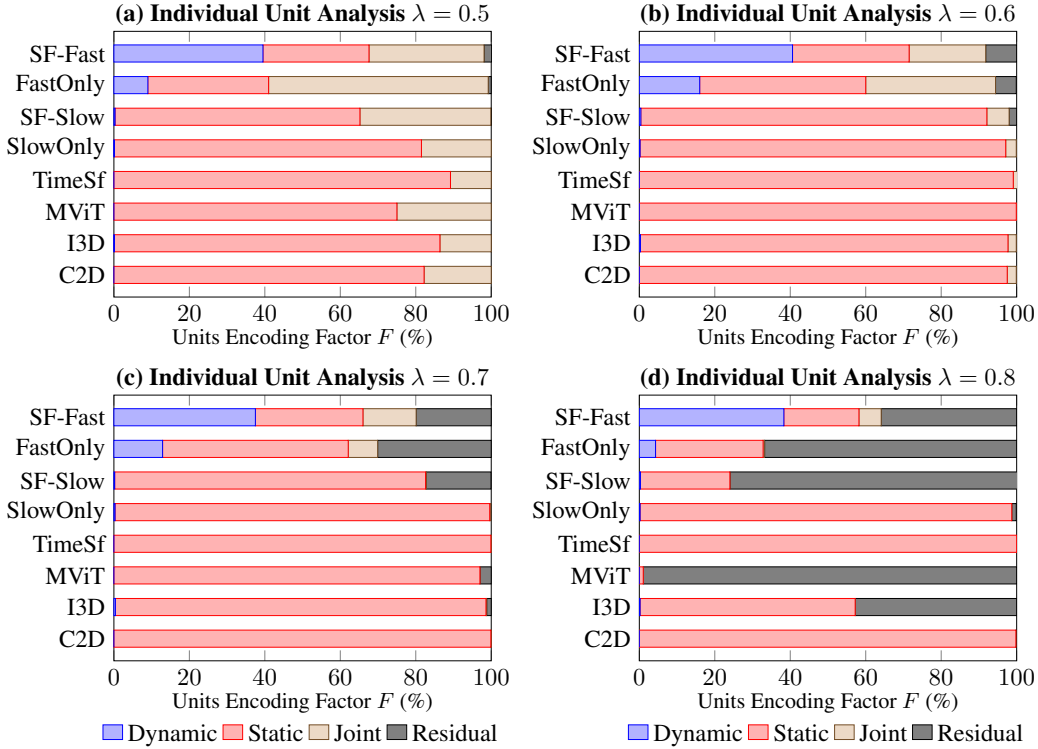
\begin{figure*}[t]
\centering
\resizebox{0.8\textwidth}{!}{
\begin{tikzpicture}
\begin{axis} [xbar stacked,
    width=7cm,
    bar width = 7pt,
    height=5.4cm,  
    xmin = 0,
    xmax = 100,
    title = \textbf{(a) Individual Unit Analysis $\lambda=0.5$},
    title style={at={(axis description cs:0.5,1.08)},anchor=north,font=\normalsize},
    x label style={at={(axis description cs:0.5,0.05)},anchor=north,font=\small},
    xlabel = Units Encoding Factor $F$ (\%),
    ytick=data,
    legend style={
            draw = none,
			area legend,
			at={(0.5,-0.2)},
			anchor=north,
			legend columns=-1},
    symbolic y coords={C2D, I3D, MViT, TimeSf, SlowOnly, SF-Slow, FastOnly, SF-Fast},
    enlarge x limits = {value = .1},
]
 



\addplot coordinates {(0,C2D) [0] (0.1,I3D) [2] (0,MViT) [0] (0,TimeSf) [0] (0.1,SlowOnly) [2] (0.3,SF-Slow) [6] (9.0,FastOnly) [23] (39.5,SF-Fast) [101]}; 
\addplot coordinates {(82.2,C2D) [1684] (86.3,I3D) [1767] (75,MViT) [576] (89.2,TimeSf) [685] (81.4,SlowOnly) [1668] (64.9,SF-Slow) [1329] (32,FastOnly) [82] (28.1,SF-Fast) [72]}; 

\addplot coordinates {(17.8,C2D) [364] (13.6,I3D) [278] (25,MViT) [192] (10.8,TimeSf) [83] (18.5,SlowOnly) [378] (34.7,SF-Slow) [710] (58.2,FastOnly) [149] (30.5,SF-Fast) [78]}; 

\addplot coordinates {(0,C2D) [0] (0.05,I3D) [1] (0,MViT) [0] (0,TimeSf) [0] (0,SlowOnly) [0] (0.14,SF-Slow) [3] (0.8,FastOnly) [2] (2,SF-Fast) [5]}; 

\end{axis}
\end{tikzpicture}

\begin{tikzpicture}
\begin{axis} [xbar stacked,
    width=7cm,
    bar width = 7pt,
    height=5.4cm,  
    xmin = 0,
    xmax = 100,
    title = \textbf{(b) Individual Unit Analysis $\lambda=0.6$},
    title style={at={(axis description cs:0.5,1.08)},anchor=north,font=\normalsize},
    x label style={at={(axis description cs:0.5,0.05)},anchor=north,font=\small},
    xlabel = Units Encoding Factor $F$ (\%),
    ytick=data,
    legend style={
            draw = none,
			area legend,
			at={(0.5,-0.2)},
			anchor=north,
			legend columns=-1},
    symbolic y coords={C2D, I3D, MViT, TimeSf, SlowOnly, SF-Slow, FastOnly, SF-Fast},
    enlarge x limits = {value = .1},
]
\addplot coordinates {(0,C2D) [0] (0.3,I3D) [2] (0,MViT) [0] (0,TimeSf) [0] (0.2,SlowOnly) [2] (0.4,SF-Slow) [6] (16,FastOnly) [23] (40.6,SF-Fast) [101]}; 
\addplot coordinates {(97.5,C2D) [1684] (97.4,I3D) [1767] (99.9,MViT) [576] (99.1,TimeSf) [685] (96.9,SlowOnly) [1668] (91.7,SF-Slow) [1329] (44,FastOnly) [82] (30.9,SF-Fast) [72]}; 

\addplot coordinates {(2.4,C2D) [364] (2.2,I3D) [278] (0.1,MViT) [192] (10.9,TimeSf) [83] (2.8,SlowOnly) [378] (5.9,SF-Slow) [710] (34.4,FastOnly) [149] (20.3,SF-Fast) [78]}; 

\addplot coordinates {(0,C2D) [0] (0.05,I3D) [1] (0,MViT) [0] (0,TimeSf) [0] (0,SlowOnly) [0] (2,SF-Slow) [3] (5.5,FastOnly) [2] (8.2,SF-Fast) [5]}; 

\end{axis}
\end{tikzpicture}}

\resizebox{0.8\textwidth}{!}{
\begin{tikzpicture}
\begin{axis} [xbar stacked,
    width=7cm,
    bar width = 7pt,
    height=5.4cm,  
    xmin = 0,
    xmax = 100,
    title = \textbf{(c) Individual Unit Analysis $\lambda=0.7$},
    title style={at={(axis description cs:0.5,1.08)},anchor=north,font=\normalsize},
    x label style={at={(axis description cs:0.5,0.05)},anchor=north,font=\small},
    xlabel = Units Encoding Factor $F$ (\%),
    ytick=data,
    legend style={
            draw = none,
			area legend,
			at={(0.5,-0.2)},
			anchor=north,
			legend columns=-1},
    symbolic y coords={C2D, I3D, MViT, TimeSf, SlowOnly, SF-Slow, FastOnly, SF-Fast},
    enlarge x limits = {value = .1},
]
\addplot coordinates {(0,C2D) [0] (0.4,I3D) [2] (0,MViT) [0] (0,TimeSf) [0] (0.3,SlowOnly) [2] (0.2,SF-Slow) [6] (12.9,FastOnly) [23] (37.5,SF-Fast) [101]}; 
\addplot coordinates {(99.9,C2D) [1684] (98.2,I3D) [1767] (97,MViT) [576] (99.9,TimeSf) [685] (99.3,SlowOnly) [1668] (82.4,SF-Slow) [1329] (49.2,FastOnly) [82] (28.5,SF-Fast) [72]}; 

\addplot coordinates {(0.05,C2D) [364] (0.2,I3D) [278] (0,MViT) [192] (0.1,TimeSf) [83] (0.3,SlowOnly) [378] (0.24,SF-Slow) [710] (7.8,FastOnly) [149] (14.1,SF-Fast) [78]}; 

\addplot coordinates {(0,C2D) [0] (1.2,I3D) [1] (3,MViT) [0] (0,TimeSf) [0] (0,SlowOnly) [0] (17.1,SF-Slow) [3] (30.1,FastOnly) [2] (19.9,SF-Fast) [5]}; 

\legend {Dynamic,Static,Joint,Residual};
\end{axis}
\end{tikzpicture}

\begin{tikzpicture}
\begin{axis} [xbar stacked,
    width=7cm,
    bar width = 7pt,
    height=5.4cm,  
    xmin = 0,
    xmax = 100,
    title = \textbf{(d) Individual Unit Analysis $\lambda=0.8$},
    title style={at={(axis description cs:0.5,1.08)},anchor=north,font=\normalsize},
    x label style={at={(axis description cs:0.5,0.05)},anchor=north,font=\small},
    xlabel = Units Encoding Factor $F$ (\%),
    ytick=data,
    legend style={
            draw = none,
			area legend,
			at={(0.5,-0.2)},
			anchor=north,
			legend columns=-1},
    symbolic y coords={C2D, I3D, MViT, TimeSf, SlowOnly, SF-Slow, FastOnly, SF-Fast},
    enlarge x limits = {value = .1},
]
\addplot coordinates {(0,C2D) [0] (0.2,I3D) [2] (0,MViT) [0] (0,TimeSf) [0] (0.2,SlowOnly) [2] (0.3,SF-Slow) [6] (4.3,FastOnly) [23] (38.3,SF-Fast) [101]}; 
\addplot coordinates {(99.8,C2D) [1684] (57.0,I3D) [1767] (1.04,MViT) [576] (100,TimeSf) [685] (98.5,SlowOnly) [1668] (23.7,SF-Slow) [1329] (28.5,FastOnly) [82] (19.9,SF-Fast) [72]}; 

\addplot coordinates {(0,C2D) [364] (0,I3D) [278] (0,MViT) [192] (0,TimeSf) [83] (0,SlowOnly) [378] (0,SF-Slow) [710] (0.4,FastOnly) [149] (5.9,SF-Fast) [78]}; 

\addplot coordinates {(0.15,C2D) [0] (42.8,I3D) [1] (98.96,MViT) [0] (0,TimeSf) [0] (1.2,SlowOnly) [0] (76.3,SF-Slow) [3] (66.8,FastOnly) [2] (35.9,SF-Fast) [5]}; 

\legend {Dynamic,Static,Joint,Residual};
\end{axis}
\end{tikzpicture}}



\caption{Estimates of the dynamic, static, joint and residual units using the unit-wise metric, (\ref{eq:ind_bias_scores_diff_b}), for the different action recognition architectures at varying values of the threshold, $\lambda$.}
\label{fig:ar_allthresholds_models}
\end{figure*}

We use standard augmentations that are found in the SlowFast repository, which include random spatial cropping and random temporal cropping, followed by resizing to $224 \times 224$. The number of frames and sampling rate for all models is $8 \times 8$ unless otherwise specified. At validation time, a single clip was spatially and temporally center cropped. All models were trained on four NVIDIA Tesla T4s. Training times for each model and dataset vary significantly. Training the SlowFast model on the Diving48 dataset takes approximately 2.5 days which is the longest training time among all considered models.


\subsubsection{SlowFast frame rates}\label{sec:ar_frame}

Figure~\ref{fig:4x16_sf} shows the static and dynamic units estimated using the layer-wise metric, (\ref{eq:biasscores}). The main paper examined architectures with a frame number and sampling rate of $8 \times 8$ while Fig.~\ref{fig:4x16_sf} shows SlowFast variants trained with a frame number and sampling rate of $4 \times 16$. It can be seen that the Fast branch injects dynamic information into the Slow branch via the fast-to-slow cross connections. The last layer of the SlowOnly model has 21.6\% units (\ie channels) encoding dynamics, while when trained jointly with the Fast branch, the SF-Slow model has 24.2\% dynamic units in the final layer. This increase of 2.6\% is similar to the one seen with sampling parameters of $8 \times 8$, which saw an increase of 3.3\% in the last layer.

\subsubsection{Varying thresholds}\label{sec:ar_vary_thresh}

Figure~\ref{fig:ar_allthresholds_models} shows the static and dynamic unit-wise analysis, (\ref{eq:ind_bias_scores_diff_b}), with varying thresholds, $\lambda$, for various action recognition architectures. The FastOnly and SlowFast-Fast models are the only ones that produce specialized dynamic units which is consistent with the findings in the main paper. Moreover, the SlowFast-Fast branch retains a significant number of dynamic units even at the higher thresholds (\eg 0.8). This pattern further shows the efficacy of using the two-stream architecture for capturing separate types of information. Note that all models produce more residual units as the threshold increases since few units have correlation coefficients in the range 0.8 to 1.

\subsubsection{Training dataset effect}\label{sec:ar_append_dataset}

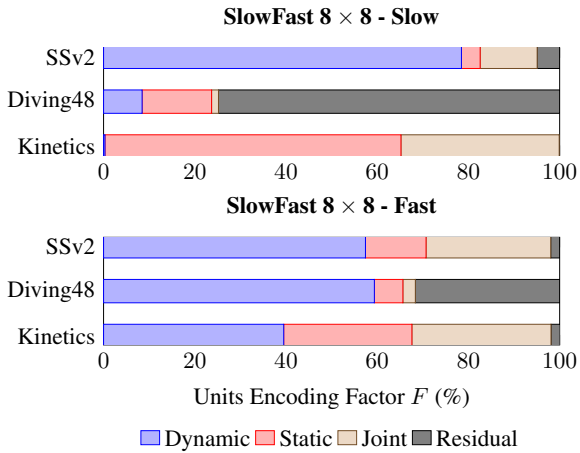
\begin{figure}
    \centering
\begin{minipage}{0.45\textwidth}
\resizebox{\textwidth}{!}{
\begin{tikzpicture}
\begin{axis} [
    width=\axisdefaultwidth,
    height=3.2cm,
    xbar stacked,
    bar width = 10pt,
    xmin = 0,
    xmax = 100,
    title = \textbf{SlowFast $\textbf{8}\times\textbf{8}$ - Slow},
    ytick=data,
    legend style={
            draw = none,
			area legend,
			at={(0.5,1.4)},
			anchor=north,
			legend columns=-1},
    symbolic y coords={Kinetics, Diving48, SSv2},
    enlarge x limits = {value = .1},
] 
\addplot coordinates { (0.3,Kinetics) (8.447265625,Diving48)  (78.466,SSv2)};
\addplot coordinates { (64.9,Kinetics) (15.234375,Diving48)  (4.1015,SSv2)};
\addplot coordinates { (34.7,Kinetics) (1.513671875,Diving48)  (12.5,SSv2)};
\addplot coordinates { (0.14,Kinetics) (74.8046875,Diving48)  (4.93164,SSv2)};
\end{axis}
\end{tikzpicture}}

\resizebox{\textwidth}{!}{
\begin{tikzpicture}
\begin{axis} [
    width=\axisdefaultwidth,
    height=3.2cm,
    xbar stacked,
    bar width = 10pt,
    xmin = 0,
    xmax = 100,
    title = \textbf{SlowFast $\textbf{8}\times\textbf{8}$ - Fast},
    xlabel = Units Encoding Factor $F$ (\%),
    ytick=data,
    legend style={
            draw = none,
			area legend,
			at={(0.5,-0.7)},
			anchor=north,
			legend columns=-1},
    symbolic y coords={Kinetics, Diving48, SSv2},
    enlarge x limits = {value = .1},
] 
\addplot coordinates { (39.5,Kinetics) (59.375,Diving48)  (57.421,SSv2)};
\addplot coordinates { (28.1,Kinetics) (6.25,Diving48)  (13.281,SSv2)};
\addplot coordinates { (30.5,Kinetics) (2.734375,Diving48)  (27.34375,SSv2)};
\addplot coordinates { (2,Kinetics) (31.640625,Diving48)  (1.95,SSv2)};

\legend {Dynamic, Static, Joint, Residual};
\end{axis}
\end{tikzpicture}}
\end{minipage}%
\caption{Analyses of static and dynamic biases of action recognition datasets using the unit-wise metric, (\ref{eq:ind_bias_scores_diff_b}), for the SlowFast architecture with the number of frames and sampling rate of $8 \times 8$. Slow and Fast represent the Slow branch and Fast branch of the SlowFast model, respectively. The estimates are based on the penultimate layer of each stream separately, before the concatenation of the representations.}
\label{fig:ar_dataset}
\end{figure}

In this section, we provide additional models trained on SSv2 and Diving48. In the main paper, we showed that SSv2 produces dynamic units and Diving48 produces residual units, while Kinetics-400 mainly produces static units. To this end, we analyse SlowFast models trained on each dataset using the unit-wise metric, (\ref{eq:ind_bias_scores_diff_b}), for $\lambda=0.5$ on the Slow and Fast branches separately; see Fig.~\ref{fig:ar_dataset}. The findings from the main paper are consistent with those seen here. Diving48 is the only dataset to produce a notable number of residual units, which suggests that there are other factors than static and dynamic that are important for classifying dives in this dataset. We leave it for future work to explore what these residual units capture. SSv2, on the other hand, yields a large number of dynamic units regardless of the architecture. Note that the Fast branch contains dynamic units regardless of the dataset, again showing the efficacy of this two-stream approach for separating static and dynamic information.

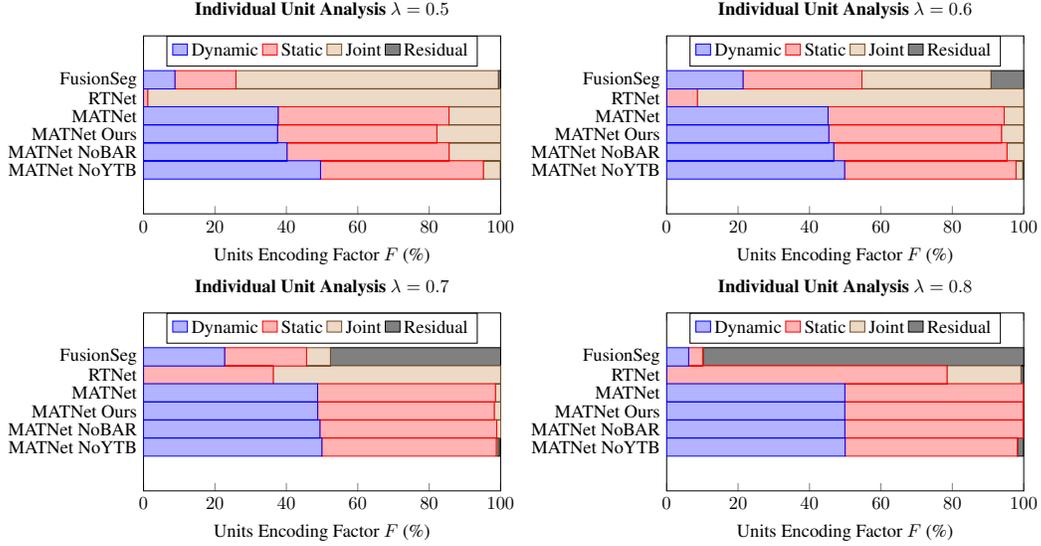
\begin{figure*}[t]
\centering
\resizebox{0.8\textwidth}{!}{
\begin{tikzpicture}
\begin{axis} [xbar stacked,
    width=\axisdefaultwidth,
    height=5cm,
    bar width = 10pt,
    xmin = 0,
    xmax = 100,
    title = \textbf{Individual Unit Analysis $\lambda=0.5$},
    xlabel = Units Encoding Factor $F$ (\%),
    ytick=data,
    legend style={
			area legend,
			at={(0.5,1)},
			anchor=north,
			legend columns=-1},
    symbolic y coords={MATNet NoYTB, MATNet NoBAR, MATNet Ours, MATNet, RTNet, FusionSeg},
    enlarge x limits = {value = .1},
    enlarge y limits={abs=24pt}
]
\addplot coordinates {(8.7890625,FusionSeg) (37.6953125,MATNet) (37.5244140625,MATNet Ours) (40.185546875,MATNet NoBAR) (49.560546875,MATNet NoYTB) (0.0,RTNet)};
\addplot coordinates {(17.08984375,FusionSeg) (47.8515625,MATNet) (44.62890625,MATNet Ours) (45.3857421875,MATNet NoBAR) (45.60546875,MATNet NoYTB) (1.171875,RTNet)};
\addplot coordinates {(73.486328125,FusionSeg) (14.453125,MATNet) (17.8466796875,MATNet Ours) (14.4287109375,MATNet NoBAR) (4.736328125,MATNet NoYTB) (98.828125,RTNet)};
\addplot coordinates {(0.634765625,FusionSeg) (0.0,MATNet) (0.0,MATNet Ours) (0.0,MATNet NoBAR) (0.0,MATNet NoYTB) (0.0,RTNet)};

\legend {Dynamic,Static,Joint,Residual};

\end{axis}
\end{tikzpicture} 
\hfill
\begin{tikzpicture}
\begin{axis} [xbar stacked,
    width=\axisdefaultwidth,
    height=5cm,
    bar width = 10pt,
    xmin = 0,
    xmax = 100,
    title = \textbf{Individual Unit Analysis $\lambda=0.6$},
    xlabel = Units Encoding Factor $F$ (\%),
    ytick=data,
    legend style={
			area legend,
			at={(0.5,1)},
			anchor=north,
			legend columns=-1},
   symbolic y coords={MATNet NoYTB, MATNet NoBAR, MATNet Ours, MATNet, RTNet, FusionSeg},
    enlarge x limits = {value = .1},
    enlarge y limits={abs=24pt}
]

\addplot coordinates {(21.38671875,FusionSeg) (45.1904296875,MATNet) (45.41015625,MATNet Ours) (46.8,MATNet NoBAR) (49.8291015625,MATNet NoYTB) (0.0,RTNet)};
\addplot coordinates {(33.30078125,FusionSeg) (49.31640625,MATNet) (48.3642578125,MATNet Ours) (48.5,MATNet NoBAR) (47.998046875,MATNet NoYTB) (8.59375,RTNet)};
\addplot coordinates {(36.1328125,FusionSeg) (5.4931640625,MATNet) (6.2,MATNet Ours) (4.7,MATNet NoBAR) (1.904296875,MATNet NoYTB) (91.40625,RTNet)};
\addplot coordinates {(9.1796875,FusionSeg) (0.0,MATNet) (0.0,MATNet Ours) (0.0,MATNet NoBAR) (0.2,MATNet NoYTB) (0.0,RTNet)};

\legend {Dynamic,Static,Joint,Residual};

\end{axis}
\end{tikzpicture}
}
\vfill
\resizebox{0.8\textwidth}{!}{
\begin{tikzpicture}
\begin{axis} [xbar stacked,
    width=\axisdefaultwidth,
    height=5cm,
    bar width = 10pt,
    xmin = 0,
    xmax = 100,
    title = \textbf{Individual Unit Analysis $\lambda=0.7$},
    xlabel = Units Encoding Factor $F$ (\%),
    ytick=data,
    legend style={
			area legend,
			at={(0.5,1)},
			anchor=north,
			legend columns=-1},
    symbolic y coords={MATNet NoYTB, MATNet NoBAR, MATNet Ours, MATNet, RTNet, FusionSeg},
    enlarge x limits = {value = .1},
    enlarge y limits={abs=24pt}
]

\addplot coordinates {(22.705078125,FusionSeg) (48.7548828125,MATNet) (48.779296875,MATNet Ours) (49.4,MATNet NoBAR) (49.951171875,MATNet NoYTB) (0.0,RTNet)};
\addplot coordinates {(22.94921875,FusionSeg) (49.8046875,MATNet) (49.4384765625,MATNet Ours) (49.5,MATNet NoBAR) (48.828125,MATNet NoYTB) (36.328125,RTNet)};
\addplot coordinates {(6.689453125,FusionSeg) (1.4404296875,MATNet) (1.8,MATNet Ours) (1.2,MATNet NoBAR) (0.5,MATNet NoYTB) (63.671875,RTNet)};
\addplot coordinates {(47.65625,FusionSeg) (0.0,MATNet) (0.0,MATNet Ours) (0.0,MATNet NoBAR) (0.6,MATNet NoYTB) (0.0,RTNet)};

\legend {Dynamic,Static,Joint,Residual};

\end{axis}
\end{tikzpicture}
\hfill
\begin{tikzpicture}
\begin{axis} [xbar stacked,
    width=\axisdefaultwidth,
    height=5cm,
    bar width = 10pt,
    xmin = 0,
    xmax = 100,
    title = \textbf{Individual Unit Analysis $\lambda=0.8$},
    xlabel = Units Encoding Factor $F$ (\%),
    ytick=data,
    legend style={
			area legend,
			at={(0.5,1)},
			anchor=north,
			legend columns=-1},
    symbolic y coords={MATNet NoYTB, MATNet NoBAR, MATNet Ours, MATNet, RTNet, FusionSeg},
    enlarge x limits = {value = .1},
    enlarge y limits={abs=24pt}
]

\addplot coordinates {(6.15234375,FusionSeg) (49.90234375,MATNet) (49.8779296875,MATNet Ours) (49.9,MATNet NoBAR) (49.9755859375,MATNet NoYTB) (0.0,RTNet)};
\addplot coordinates {(3.90625,FusionSeg) (49.951171875,MATNet)  (49.90234375,MATNet Ours) (49.9,MATNet NoBAR) (48.2421875,MATNet NoYTB) (78.515625,RTNet)};
\addplot coordinates {(0.29296875,FusionSeg) (0.146484375,MATNet) (0.2,MATNet Ours) (0.17,MATNet NoBAR) (0.1,MATNet NoYTB) (20.703125,RTNet)};
\addplot coordinates {(89.6484375,FusionSeg) (0.0,MATNet) (0.0,MATNet Ours) (0.0,MATNet NoBAR) (1.6,MATNet NoYTB) (0.7,RTNet)};

\legend {Dynamic,Static,Joint,Residual};

\end{axis}
\end{tikzpicture}
}
\caption{Estimates of the dynamic, static, joint and residual units using our metric for unit-wise analysis, (\ref{eq:ind_bias_scores_diff_b}), for the different VOS models at various settings of the threshold, $\lambda$.}
\label{fig:vos_allthresholds}
\end{figure*}

\begin{figure*}[t]
\centering
\resizebox{0.8\textwidth}{!}{
\begin{tikzpicture}
\begin{axis} [xbar stacked,
    width=\axisdefaultwidth,
    height=5cm,
    bar width = 10pt,
    xmin = 0,
    xmax = 100,
    title = \textbf{Fusion Layer 5},
    xlabel = Units Encoding Factor $F$ (\%),
    ytick=data,
    legend style={
			area legend,
			at={(0.5,1.)},
			anchor=north,
			legend columns=-1},
    symbolic y coords={MATNet NoYTB, MATNet NoBAR, MATNet Ours, MATNet, RTNet, FusionSeg},
    enlarge x limits = {value = .1},
    enlarge y limits={abs=24pt}
]

\addplot coordinates {(8.7890625,FusionSeg) (37.6953125,MATNet) (37.5244140625,MATNet Ours) (40.185546875,MATNet NoBAR) (49.560546875,MATNet NoYTB) (0.0,RTNet)};
\addplot coordinates {(17.08984375,FusionSeg) (47.8515625,MATNet)  (44.62890625,MATNet Ours) (45.3857421875,MATNet NoBAR) (45.60546875,MATNet NoYTB) (1.171875,RTNet)};
\addplot coordinates {(73.486328125,FusionSeg) (14.453125,MATNet) (17.8466796875,MATNet Ours) (14.4287109375,MATNet NoBAR) (4.736328125,MATNet NoYTB) (98.828125,RTNet)};
\addplot coordinates {(0.634765625,FusionSeg) (0.0,MATNet) (0.0,MATNet Ours) (0.0,MATNet NoBAR) (0.09,MATNet NoYTB) (0.0,RTNet)};

\legend {Dynamic,Static,Joint,Residual};

\end{axis}
\end{tikzpicture} 
\hfill
\begin{tikzpicture}
\begin{axis} [xbar stacked,
    width=\axisdefaultwidth,
    height=5cm,
    bar width = 10pt,
    xmin = 0,
    xmax = 100,
    title = \textbf{Fusion Layer 4},
    xlabel = Units Encoding Factor $F$ (\%),
    ytick=data,
    legend style={
			area legend,
			at={(0.5,1)},
			anchor=north,
			legend columns=-1},
    symbolic y coords={MATNet NoYTB, MATNet NoBAR, MATNet Ours, MATNet, RTNet, FusionSeg},
    enlarge x limits = {value = .1},
    enlarge y limits={abs=24pt}
]

\addplot coordinates {(28.6,MATNet) (33.7890625,MATNet Ours) (30.3,MATNet NoBAR) (46.044921875,MATNet NoYTB) (0.0,RTNet)};
\addplot coordinates {(40.3,MATNet) (37.40234375,MATNet Ours) (39.0,MATNet NoBAR) (32.080078125,MATNet NoYTB) (65.6,RTNet)};
\addplot coordinates {(31.1,MATNet) (28.80859375,MATNet Ours) (30.7,MATNet NoBAR) (9.716796875,MATNet NoYTB) (34.4,RTNet)};
\addplot coordinates {(0.0,MATNet) (0.0,MATNet Ours) (0.0,MATNet NoBAR) (12.158203125,MATNet NoYTB) (0.0,RTNet)};

\legend {Dynamic,Static,Joint,Residual};

\end{axis}
\end{tikzpicture}
}
\vfill
\resizebox{0.8\textwidth}{!}{
\begin{tikzpicture}
\begin{axis} [xbar stacked,
    width=\axisdefaultwidth,
    height=5cm,
    bar width = 10pt,
    xmin = 0,
    xmax = 100,
    title = \textbf{Fusion Layer 3},
    xlabel = Units Encoding Factor $F$ (\%),
    ytick=data,
    legend style={
			area legend,
			at={(0.5,1)},
			anchor=north,
			legend columns=-1},
    symbolic y coords={MATNet NoYTB, MATNet NoBAR, MATNet Ours, MATNet, RTNet, FusionSeg},
    enlarge x limits = {value = .1},
    enlarge y limits={abs=24pt}
]
\addplot coordinates {(38.3,MATNet) (33.69140625,MATNet Ours) (33.3,MATNet NoBAR) (30.95703125,MATNet NoYTB) (3.9,RTNet)};
\addplot coordinates {(43.7,MATNet) (46.6796875,MATNet Ours) (44.9,MATNet NoBAR) (47.75390625,MATNet NoYTB) (64.1,RTNet)};
\addplot coordinates {(18.1,MATNet) (19.62890625,MATNet Ours) (21.8,MATNet NoBAR) (20.99609375,MATNet NoYTB) (26.6,RTNet)};
\addplot coordinates {(0.0,MATNet) (0.0,MATNet Ours) (0.0,MATNet NoBAR) (0.29296875,MATNet NoYTB) (5.5,RTNet)};

\legend {Dynamic,Static,Joint,Residual};

\end{axis}
\end{tikzpicture}
\begin{tikzpicture}
\begin{axis} [xbar stacked,
     width=\axisdefaultwidth,
    height=5cm,
    bar width = 10pt,
    xmin = 0,
    xmax = 100,
    title = \textbf{Fusion Layer 2},
    xlabel = Units Encoding Factor $F$ (\%),
    ytick=data,
    legend style={
			area legend,
			at={(0.5,1)},
			anchor=north,
			legend columns=-1},
    symbolic y coords={MATNet NoYTB, MATNet NoBAR, MATNet Ours, MATNet, RTNet, FusionSeg},
    enlarge x limits = {value = .1},
    enlarge y limits={abs=24pt}
]
\addplot coordinates {(32.0,MATNet) (29.8828125,MATNet Ours) (31.1,MATNet NoBAR) (20.5078125,MATNet NoYTB) (6.3,RTNet) (32.421875,FusionSeg)};
\addplot coordinates {(45.9,MATNet) (46.2890625,MATNet Ours) (44.1,MATNet NoBAR) (52.5390625,MATNet NoYTB) (59.4,RTNet) (12.109375,FusionSeg)};
\addplot coordinates {(22.1,MATNet) (23.828125,MATNet Ours) (24.8,MATNet NoBAR) (26.7578125,MATNet NoYTB) (21.9,RTNet) (54.296875,FusionSeg)};
\addplot coordinates {(0.0,MATNet) (0.0,MATNet Ours) (0.0,MATNet NoBAR) (0.1953125,MATNet NoYTB) (12.5,RTNet) (1.171875,FusionSeg)};

\legend {Dynamic,Static,Joint,Residual};

\end{axis}
\end{tikzpicture}
}
\caption{Estimates of the dynamic, static, joint and residual units using our metric for unit-wise analysis, (\ref{eq:ind_bias_scores_diff_b}), $\lambda=0.5$ comparing MATNet variants and RTNet at different fusion layers. Our modified version of FusionSeg has only two fusion layers (\ie fusion layer two and five) to be compared.}
\label{fig:vos_alllayers}
\end{figure*}
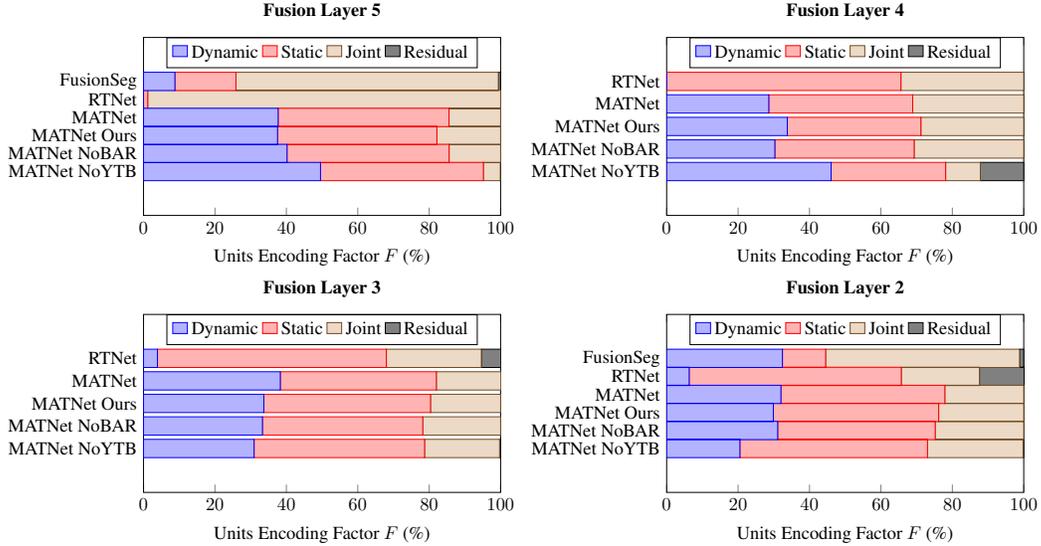

\subsection{Video object segmentation}\label{sec:vos}

In this section, we provide details for video object segmentation. We begin by presenting implementation details for all models evaluated in the main submission. Subsequently, we provide a series of supplementary experiments where we consider the effect of architectures on static vs. dynamic encoding. For each architecture we (a) consider the effect of the threshold, $\lambda$, in our unit-wise metric, (\ref{eq:ind_bias_scores_diff_b}), (b) examine all fusion layers and (c) present supporting experiments on a dataset that especially emphasizes the importance of motion in segmentation, as the objects of interest are camouflaged in single frames, MoCA~\cite{lamdouar2020betrayed}. Finally, we demonstrate the individual unit analysis on different VOS datasets with varying thresholds, $\lambda$, to provide additional confirmation of our final conclusions.

\subsubsection{Implementation details}\label{sec:impVOS}
In this subsection, we describe the implementation details for FusionSeg~\cite{jain2017fusionseg} modified version, MATNet~\cite{zhou2020motion} variants and the evaluation on MoCA~\cite{lamdouar2020betrayed}. In our modified version of FusionSeg we follow the original in using a ResNet-101~\cite{he2016deep} backbone with five stages, the first being the early convolutional layers and the rest being four ResNet-like stages. However, unlike the original work, we apply the fusion between both motion and appearance features on the intermediate representations at stages two and five. As explained in the main submission, we make this adjustment to allow for comparison with MATNet and RTNet that perform fusion on the intermediate representations. We use $1\times1$ convolutional layers for the fusion, which take concatenated features and output 256 and 2048 feature channels at stages two and five, respectively. Similar to the original approach, the segmentation decoder uses atrous spatial pyramid pooling (ASPP), but we also use an encoder-decoder architecture~\cite{chen2018encoder}. Specifically, we concatenate the features extracted from the fusion of stage two and ASPP features to produce the final segmentation mask. 

Our FusionSeg model is trained with a batch size of eight, using SGD with learning rate 0.001, along with a momentum of 0.9, and weight decay $1\times10^{-4}$. Additionally, we use a ``poly'' learning rate policy using a power of 0.9. We use random scaling with scale randomly sampled between (0.5, 2.0), random cropping with size $513 \times 513$ and random horizontal flipping for data augmentation. We do not perform pretraining for the motion stream, unlike what was proposed in the original paper. We make this choice because we focus on training the joint model directly on three different datasets to assess their effect.

MATNet variants are trained on a multi-GPU (with two GPUs 1080 Ti) machine with batch size six (unlike the original MATNet that used batch two and no multi-GPU training), the rest of the training hyperparameters follow the original work~\cite{zhou2020motion}. We denote the original model provided by the authors without finetuning or training on our side as ``MATNet'', while reproduction of MATNet with training on multi-GPU and a batch size of six as ``MATNet Ours''. We train a MATNet variant without boundary aware refinement (BAR)  modules that we call ``MATNet NoBAR'', where we remove all BAR modules and the boundary loss. We also experiment with another MATNet variant that does not train on additional YouTubeVOS data~\cite{xu2018youtube}, unlike the originally proposed model, we call this version ``MATNet NoYTB''. We specifically introduce MATNet Ours to provide a MATNet variant that is directly comparable to other variants that we introduce (\ie MATNet NoBAR and MATNet NoYTB), as it has the same training paradigm, unlike the original MATNet~\cite{zhou2020motion}.
We evaluate the static and dynamic units for these variants to investigate the reason behind the increased dynamic units with respect to other models considered (\ie FusionSeg and RTNet). We make no modifications to RTNet and use the public version provided by the authors \cite{ren2021reciprocal}. They provide a model with a ResNext50 backbone, which we denote as ``RTNet'' throughout the paper. For all architectures, we use RAFT~\cite{teed2020raft} to supply the optical flow estimates used for sampling of static and dynamic pairs on the stylized DAVIS16 validation dataset.

Finally, we describe the evaluation details on the Moving Camouflaged Animals dataset (MoCA)~\cite{lamdouar2020betrayed}. We follow previous work by removing videos that contain no predominant target locomotion, which produces a subset of 88 videos for evaluation~\cite{yang2021selfsupervised}. We evaluate using mean intersection over union and success rate with varying IoU thresholds, $\tau$, ranging from 0.5 to 0.9. We evaluate our modified FusionSeg, RTNet provided by the original work, and our MATNet variants on MoCA. The original MATNet evaluation on MoCA is reported in previous work~\cite{yang2021selfsupervised}. It is worth noting, that the original MATNet used data augmentation as horizontal flipping during the inference and averaged predictions from the original and flipped versions. To ensure fair comparison with RTNet and FusionSeg on MoCA we disable the data augmentation during inference when reporting on MoCA.

\subsubsection{Architectural effect}\label{sec:vos_arch_effect_appendix}

Figure~\ref{fig:vos_allthresholds} shows the unit-wise analysis, (\ref{eq:ind_bias_scores_diff_b}), on all studied VOS architectures
with various settings of the threshold, $\lambda$, for the late fusion layer (\ie fusion layer five). We vary the threshold, $\lambda$, between $0.5$ to $0.8$. It is seen that the off-the-shelf MATNet~\cite{zhou2020motion} consistently contains more dynamic units than both RTNet~\cite{ren2021reciprocal} and our modified FusionSeg~\cite{jain2017fusionseg}. For RTNet, increased values of $\lambda$ yield an increased number of static units at the expense of joint units, while the number of dynamic units always remain small. For FusionSeg, both dynamic and static units initially increase at the expense of joint units as $\lambda$ increases; however, at the highest value of $\lambda$ the majority of units become residuals. The pattern of decreased numbers of joint units with increased values of $\lambda$ arises because the requirement for units to be judged as jointly encoding becomes increasingly stringent; see (\ref{eq:ind_bias_scores_diff_b}).

\begin{table*}[t!]
    \centering
\begin{tabu}{@{}lccccccc}
\tabucline[1pt]{-}
\multirow{2}{*}{Method} & \multirow{2}{*}{mIoU} & \multicolumn{6}{c}{Success Rate} \\ \cmidrule(r{2pt}){3-8}
&   & $\tau=0.5$ & $\tau=0.6$ & $\tau=0.7$& $\tau=0.8$ &  $\tau=0.9$ &  $SR_{\text{Mean}}$\\ \tabucline[1pt]{-}
FusionSeg~\cite{jain2017fusionseg} Modified & 42.3 & 47.9 & 43.6 & 35.9 & 24.2 & 9.4 & 39.2\\
RTNet~\cite{ren2021reciprocal} & 60.7 & 67.9 & 62.4 & 53.6 & 43.4 & 23.9 & 50.2 \\ 
MATNet~\cite{zhou2020motion} & 64.2 & 71.2 & 67.0 & 59.9 & \textbf{49.2} & 24.6 & 54.4 \\\hline
MATNet Ours & \textbf{67.3} & \textbf{75.9} & \textbf{70.8} & \textbf{61.9} & 48.6 	& \textbf{26.0} & \textbf{56.6}\\
MATNet NoBAR & 65.1 & 73.6 & 68.0 & 58.9 & 44.7 & 21.5 &  53.3\\
MATNet NoYTB & 54.7 & 59.9 & 53.5 & 44.0 & 31.0 & 13.4 & 40.3 \\ \hline
\tabucline[1pt]{-}
\end{tabu}
    \caption{Evaluation of VOS models on MoCA~\cite{lamdouar2020betrayed}. Success rates are reported using different IoU thresholds, $\tau$. FusionSeg, RTNet and MATNet are the same versions reported in the main submission and can be compared directly. The three MATNet variants (MATNet Ours, MATNet NoBAR and MATNet NoYTB) are trained on our side, while MATNet~\cite{zhou2020motion} is the original without our training; see Sec.~\ref{sec:impVOS}. Thus, while the three variants trained on our side are directly comparable, they cannot be compared directly with the original MATNet.} 
    \label{tab:moca_vs_davis}
\end{table*}
\begin{figure}[t!]
\centering
\resizebox{0.4\textwidth}{!}{
	\begin{tikzpicture}\ref{jointdynamic_legend}
    \begin{axis}[
       line width=1.0,
        title style={at={(axis description cs:0.5,1.15)},anchor=north,font=\normalsize},
        ylabel={mIoU on MoCA (\%)},
        xlabel={Relative Joint/Dynamic (\%)},
        ymin=40, ymax=70,
        xmin=0, xmax=50,
        ytick={40,50,60, 70},
        xtick={0, 10, 20, 30, 40, 50},
        x tick label style={font=\small},
        y tick label style={font=\normalsize},
        x label style={at={(axis description cs:0.5,0.03)},anchor=north,font=\small},
        y label style={at={(axis description cs:0.12,.5)},anchor=south,font=\normalsize},
        width=8cm,
        height=7cm,        
        ymajorgrids=false,
        xmajorgrids=false,
        major grid style={dotted,green!20!black},
        legend style={
         nodes={scale=0.9, transform shape},
         cells={anchor=west},
         legend style={at={(3.8,0.25)},anchor=south}, font =\small},
         legend entries={[black]MATNet Ours, [black]MATNet NoBAR,[black]MATNet NoYTB},
        legend to name=jointdynamic_legend,
    ]

    \addplot[only marks,mark size=3pt, color=cyan,mark=*] coordinates {(47.5, 66.9)};
    \addplot[only marks,mark size=3pt, color=red,mark=+] coordinates {(35.8, 65.1)};
    \addplot[only marks,mark size=3pt, color=blue,mark=o] coordinates {(9.5, 54.7)};
	    
    \end{axis}
\end{tikzpicture}}

\caption{The relative joint/dynamic units in the final fusion layer compared with the mean intersection over union on MoCA for dynamically biased models (\ie MATNet variants).}
\label{fig:jointreldyn}
\end{figure}
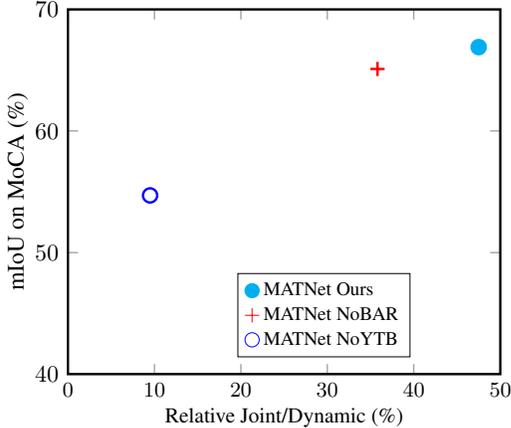
In further comparing the different MATNet variants (including the variant that lacks boundary aware refinement modules and the variant that lacks additional YouTubeVOS training) to FusionSeg and RTNet, it is seen that the proportion of dynamic units remains higher. This result suggests that the source behind the increase in the number of dynamic units in MATNet is the motion-to-appearance cross connections, rather than additional data or boundary refinement. It also shows for MATNet variants that with higher settings of the threshold, $\lambda$, the joint encoding units decrease and the specialized static/dynamic units increase.
Figure~\ref{fig:vos_alllayers} shows the comparison among different VOS architectures on all fusion layers (\ie fusion layers two, three, four and five). It is seen that in fusion layers three, four and five, MATNet consistently has more dynamic units compared to FusionSeg and RTNet, which instead tend to have more jointly encoding units. The same pattern was shown in the main submission, albeit only for the final fusion layer. It is also seen that MATNet tends to balance between the specialized units of both static and dynamic factors. 

Interestingly, however, in fusion layer two, FusionSeg appears on par with MATNet in terms of dynamic units, while it has fewer static units and more joint units. In comparison, RTNet tends to have the most unbalanced units of all three models, which are skewed toward the jointly encoding units in the late fusion layers (\ie three, four and five). This pattern confirms that models with less ability to capture dynamics in the late fusion layers (\ie FusionSeg and RTNet), generally tend to favour jointly encoding units over specialized units, and have less balance between both static and dynamic units. Thus, over all fusion layers and thresholds MATNet consistently has more balance between dynamic and static units and generally more dynamic units than other models, making models with cross connections that are not pretrained on saliency segmentation datasets one of the best to capture dynamics.

To further support the conclusion that MATNet's architecture provides the best ability to capture dynamics, we evaluate on a downstream task that requires capturing dynamics (\ie segmenting moving camouflaged animals). In particular, we evaluate all models on the MoCA subset reported in previous work~\cite{yang2021selfsupervised} and show results in Table~\ref{tab:moca_vs_davis}. 
It is seen that the original MATNet and MATNet Ours both outperform RTNet and FusionSeg when motion is key to segmentation (MoCA). 

\begin{figure*}[t!]
\centering
\resizebox{0.8\textwidth}{!}{
\begin{tikzpicture}
\begin{axis} [xbar stacked,
    width=\axisdefaultwidth,
    height=4cm,
    bar width = 10pt,
    xmin = 0,
    xmax = 100,
    title = \textbf{Fusion Layer 5 $\lambda=0.5$},
    xlabel = Units Encoding Factor $F$ (\%),
    ytick=data,
    legend style={
			area legend,
			at={(0.5,1)},
			anchor=north,
			legend columns=-1},
     symbolic y coords={TAO-VOS, ImageNetVID, DAVIS},
    enlarge x limits = {value = .1},
    enlarge y limits={abs=24pt}
]%
\addplot coordinates {(49.70703125,TAO-VOS) (0.9765625,ImageNetVID) (8.7890625,DAVIS)};
\addplot coordinates {(18.5546875,TAO-VOS) (20.654296875,ImageNetVID) (17.08984375,DAVIS)};
\addplot coordinates {(22.021484375,TAO-VOS) (78.173828125,ImageNetVID) (73.486328125,DAVIS)};
\addplot coordinates {(9.716796875,TAO-VOS) (0.1953125,ImageNetVID) (0.634765625,DAVIS)};

\legend {Dynamic,Static,Joint,Residual};

\end{axis}
\end{tikzpicture} 
\hfill
\begin{tikzpicture}
\begin{axis} [xbar stacked,
    width=\axisdefaultwidth,
    height=4cm,
    bar width = 10pt,
    xmin = 0,
    xmax = 100,
    title = \textbf{Fusion Layer 5 $\lambda=0.6$},
    xlabel = Units Encoding Factor $F$ (\%),
    ytick=data,
    legend style={
			area legend,
			at={(0.5,1)},
			anchor=north,
			legend columns=-1},
     symbolic y coords={TAO-VOS, ImageNetVID, DAVIS},
    enlarge x limits = {value = .1},
    enlarge y limits={abs=24pt}
]
\addplot coordinates {(42.578125,TAO-VOS) (4.00390625,ImageNetVID) (21.38671875,DAVIS)};
\addplot coordinates {(21.044921875,TAO-VOS) (44.140625,ImageNetVID) (33.30078125,DAVIS)};
\addplot coordinates {(7.32421875,TAO-VOS) (49.365234375,ImageNetVID) (36.1328125,DAVIS)};
\addplot coordinates {(29.052734375,TAO-VOS) (2.490234375,ImageNetVID) (9.1796875,DAVIS)};

\legend {Dynamic,Static,Joint,None};

\end{axis}
\end{tikzpicture}
}
\vfill
\resizebox{0.8\textwidth}{!}{
\begin{tikzpicture}
\begin{axis} [xbar stacked,
    width=\axisdefaultwidth,
    height=4cm,
    bar width = 10pt,
    xmin = 0,
    xmax = 100,
    title = \textbf{Fusion Layer 5 $\lambda=0.7$},
    xlabel = Units Encoding Factor $F$ (\%),
    ytick=data,
    legend style={
			area legend,
			at={(0.5,1)},
			anchor=north,
			legend columns=-1},
     symbolic y coords={TAO-VOS, ImageNetVID, DAVIS},
    enlarge x limits = {value = .1},
    enlarge y limits={abs=24pt}
]
\addplot coordinates {(23.14453125,TAO-VOS) (5.908203125,ImageNetVID) (22.705078125,DAVIS)};
\addplot coordinates {(13.525390625,TAO-VOS) (59.716796875,ImageNetVID) (22.94921875,DAVIS)};
\addplot coordinates {(1.171875,TAO-VOS) (17.431640625,ImageNetVID) (6.689453125,DAVIS)};
\addplot coordinates {(62.158203125,TAO-VOS) (16.943359375,ImageNetVID) (47.65625,DAVIS)};

\legend {Dynamic,Static,Joint,None};

\end{axis}
\end{tikzpicture}
\hfill
\begin{tikzpicture}
\begin{axis} [xbar stacked,
    width=\axisdefaultwidth,
    height=4cm,
    bar width = 10pt,
    xmin = 0,
    xmax = 100,
    title = \textbf{Fusion Layer 5 $\lambda=0.8$},
    xlabel = Units Encoding Factor $F$ (\%),
    ytick=data,
    legend style={
			area legend,
			at={(0.5,1)},
			anchor=north,
			legend columns=-1},
     symbolic y coords={TAO-VOS, ImageNetVID, DAVIS},
    enlarge x limits = {value = .1},
    enlarge y limits={abs=24pt}
]
\addplot coordinates {(5.56640625,TAO-VOS) (2.490234375,ImageNetVID) (6.15234375,DAVIS)};
\addplot coordinates {(4.296875,TAO-VOS) (34.033203125,ImageNetVID) (3.90625,DAVIS)};
\addplot coordinates {(0.048828125,TAO-VOS) (1.171875,ImageNetVID) (0.29296875,DAVIS)};
\addplot coordinates {(90.087890625,TAO-VOS) (62.3046875,ImageNetVID) (89.6484375,DAVIS)};

\legend {Dynamic,Static,Joint,Residual};

\end{axis}
\end{tikzpicture}
}
\caption{Estimates of the dynamic, static, joint and residual units using the unit-wise metric, (\ref{eq:ind_bias_scores_diff_b}), for the different VOS datasets at various settings of the threshold, $\lambda$, for fusion layer five.}
\label{fig:vos_allthresholds_dataset}
\end{figure*}
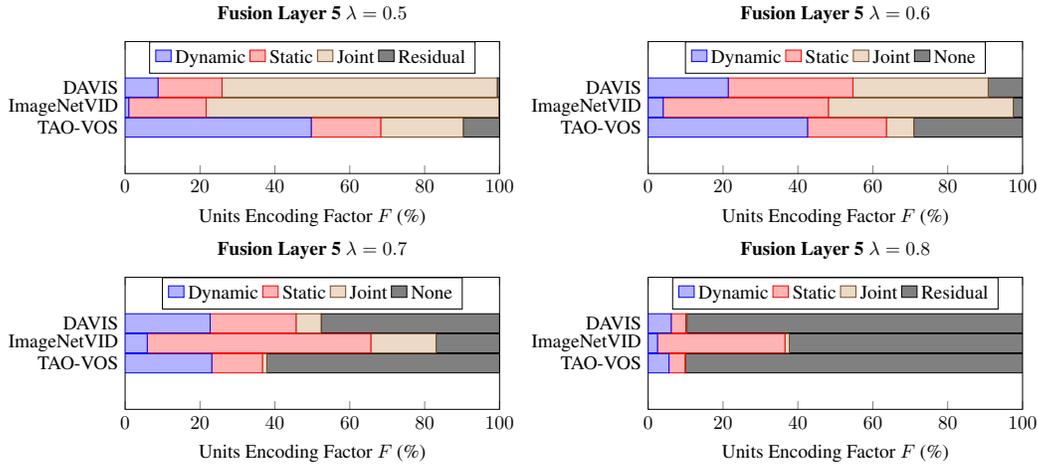

\begin{figure*}[t!]
\centering
\resizebox{0.8\textwidth}{!}{
\begin{tikzpicture}
\begin{axis} [xbar stacked,
    width=\axisdefaultwidth,
    height=4cm,
    bar width = 10pt,
    xmin = 0,
    xmax = 100,
    title = \textbf{Fusion Layer 2 $\lambda=0.5$},
    xlabel = Units Encoding Factor $F$ (\%),
    ytick=data,
    legend style={
			area legend,
			at={(0.5,1)},
			anchor=north,
			legend columns=-1},
     symbolic y coords={TAO-VOS, ImageNetVID, DAVIS},
    enlarge x limits = {value = .1},
    enlarge y limits={abs=24pt}
]
\addplot coordinates { (32.421875,DAVIS) (60.15625,ImageNetVID)  (70.703125,TAO-VOS)};
\addplot coordinates { (12.109375,DAVIS) (8.203125,ImageNetVID)  (3.515625,TAO-VOS)};
\addplot coordinates { (54.296875,DAVIS) (30.46875,ImageNetVID)  (23.828125,TAO-VOS)};
\addplot coordinates { (1.171875,DAVIS) (1.171875,ImageNetVID)  (1.953125,TAO-VOS)};

\legend {Dynamic,Static,Joint,Residual};

\end{axis}
\end{tikzpicture} 
\hfill
\begin{tikzpicture}
\begin{axis} [xbar stacked,
    width=\axisdefaultwidth,
    height=4cm,
    bar width = 10pt,
    xmin = 0,
    xmax = 100,
    title = \textbf{Fusion Layer 2 $\lambda=0.6$},
    xlabel = Units Encoding Factor $F$ (\%),
    ytick=data,
    legend style={
			area legend,
			at={(0.5,1)},
			anchor=north,
			legend columns=-1},
     symbolic y coords={TAO-VOS, ImageNetVID, DAVIS},
    enlarge x limits = {value = .1},
    enlarge y limits={abs=24pt}
]
\addplot coordinates {(83.984375,TAO-VOS) (76.171875,ImageNetVID) (55.46875,DAVIS)};
\addplot coordinates {(3.5,TAO-VOS) (7.4,ImageNetVID) (14.0625,DAVIS)};
\addplot coordinates {(7.0,TAO-VOS) (8.2,ImageNetVID) (24.21875,DAVIS)};
\addplot coordinates {(5.5,TAO-VOS) (8.2,ImageNetVID) (6.25,DAVIS)};

\legend {Dynamic,Static,Joint,None};

\end{axis}
\end{tikzpicture}
}
\vfill
\resizebox{0.8\textwidth}{!}{
\hfill
\begin{tikzpicture}
\begin{axis} [xbar stacked,
    width=\axisdefaultwidth,
    height=4cm,
    bar width = 10pt,
    xmin = 0,
    xmax = 100,
    title = \textbf{Fusion Layer 2 $\lambda=0.7$},
    xlabel = Units Encoding Factor $F$ (\%),
    ytick=data,
    legend style={
			area legend,
			at={(0.5,1)},
			anchor=north,
			legend columns=-1},
     symbolic y coords={TAO-VOS, ImageNetVID, DAVIS},
    enlarge x limits = {value = .1},
    enlarge y limits={abs=24pt}
]
\addplot coordinates {(79.0,TAO-VOS) (72.65625,ImageNetVID) (65.625,DAVIS)};
\addplot coordinates {(3.9,TAO-VOS) (3.9,ImageNetVID) (10.15625,DAVIS)};
\addplot coordinates {(0.7,TAO-VOS) (0.0,ImageNetVID) (2.7,DAVIS)};
\addplot coordinates {(16.4,TAO-VOS) (22.7,ImageNetVID) (21.484375,DAVIS)};

\legend {Dynamic,Static,Joint,None};

\end{axis}
\end{tikzpicture}
\hfill
\begin{tikzpicture}
\begin{axis} [xbar stacked,
    width=\axisdefaultwidth,
    height=4cm,
    bar width = 10pt,
    xmin = 0,
    xmax = 100,
    title = \textbf{Fusion Layer 2 $\lambda=0.8$},
    xlabel = Units Encoding Factor $F$ (\%),
    ytick=data,
    legend style={
			area legend,
			at={(0.5,1)},
			anchor=north,
			legend columns=-1},
     symbolic y coords={TAO-VOS, ImageNetVID, DAVIS},
    enlarge x limits = {value = .1},
    enlarge y limits={abs=24pt}
]

\addplot coordinates {(66.015625,TAO-VOS) (55.859375,ImageNetVID) (50.78125,DAVIS)};
\addplot coordinates {(0.0,TAO-VOS) (1.2,ImageNetVID) (5.078125,DAVIS)};
\addplot coordinates {(0.0,TAO-VOS) (0.0,ImageNetVID) (0.0,DAVIS)};
\addplot coordinates {(34.0,TAO-VOS) (43.0,ImageNetVID) (44.140625,DAVIS)};

\legend {Dynamic,Static,Joint,Residual};

\end{axis}
\end{tikzpicture}
}
\caption{Estimates of the dynamic, static, joint and residual units using the unit-wise metric, (\ref{eq:ind_bias_scores_diff_b}), for the different VOS datasets at various settings of the threshold, $\lambda$, for fusion layer two.}
\label{fig:vos_allthresholds_dataset_early}
\end{figure*}
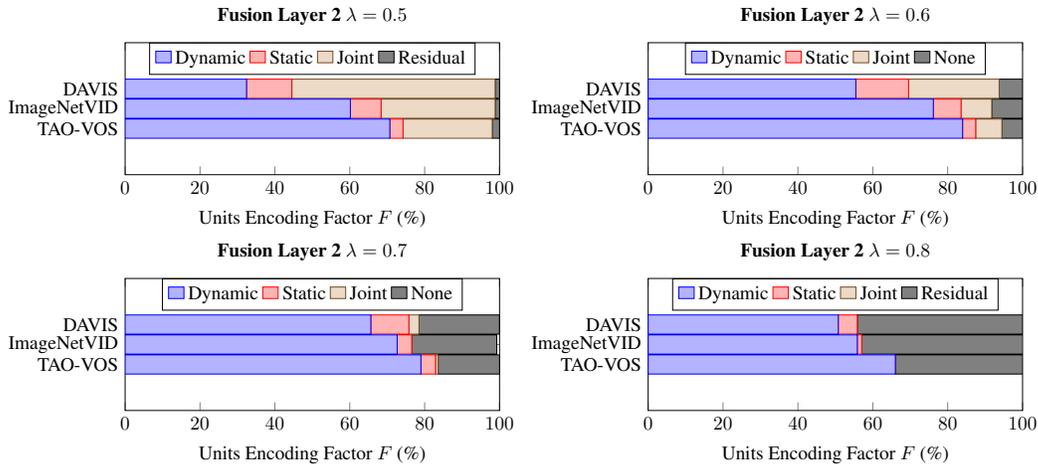

We now further pursue the main driving factors behind MATNet's improved performance on MoCA over alternative state-of-the-art models. Figure~\ref{fig:jointreldyn} shows the mean intersection over union on MoCA with respect to the relative joint to dynamic units in the final fusion layer in different MATNet variants. For this experiment, we consider only the MATNet variants trained on our side (MATNet Ours, MATNet NoBAR and MATNet NoYTB), as they are directly comparable, unlike the original MATNet; see Sec.~\ref{sec:impVOS}.
The best MATNet variant on MoCA is the one trained with additional YouTubeVOS data and using the boundary aware refinement modules with auxiliary boundary losses (\ie MATNet Ours). Interestingly, having more dynamic units along with maintaining a relative number of joint to dynamic units above a certain threshold improves MoCA performance. All MATNet variants generally have more dynamic units in their fusion layers than the rest of the VOS models. 
This suggests the driving reasons behind the state-of-the-art performance of MATNet on MoCA encompasses two main choices: (i) the inclusion of cross connections and (ii) additional training with YouTubeVOS. 



\subsubsection{Training dataset effect}


In this section, we conduct additional experiments for understanding the training dataset effect on our modified version of FusionSeg~\cite{jain2017fusionseg} by augmenting the results shown in the main submission by varying $\lambda$. Figures~\ref{fig:vos_allthresholds_dataset} and~\ref{fig:vos_allthresholds_dataset_early} show results obtained with our unit-wise metric, (\ref{eq:ind_bias_scores_diff_b}), for both fusion layers five and two. It is seen that TAO-VOS yields more specialized dynamic units than ImageNet VID with all thresholds, $\lambda$. It is also seen that TAO-VOS in fusion layer five yields more specialized dynamic units with respect to DAVIS16 on thresholds $\lambda=\{0.5, 0.6\}$ and then starts to be on-par with DAVIS16 at higher thresholds. In contrast, in fusion layer two TAO-VOS yields more dynamic units than DAVIS16. Consistently, it is further seen that there are a higher number of residual units resulting from TAO-VOS than the other two datasets in fusion layer five for thresholds $\lambda=\{0.5,0.6,0.7\}$. These results indicate that there are also other factors beyond static and dynamic factors that are captured when training on TAO-VOS. We leave it for future work to explore what these residual units capture.

\subsection{Neuron mask removal}
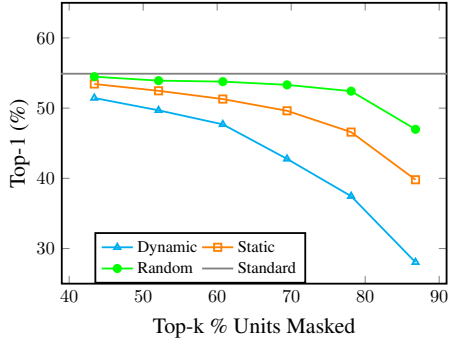
\begin{figure} [t]
	\begin{center}
     \centering 
     \resizebox{0.35\textwidth}{!}{
\begin{tikzpicture} 
                 \begin{axis}[
                 line width=1.0,
                 title style={at={(axis description cs:0.5,1.2)},anchor=north,font=\normalsize},
                 xlabel={Top-k \% Units Masked},
                 ylabel={Top-1 (\%)},
                 xmin=39, xmax=91,
                 ymin=25, ymax=65,
                 xtick={40,50,60,70,80,90},
                 x tick label style={font=\normalsize, rotate=0, anchor=north},
                 y tick label style={font=\normalsize},
                 x label style={at={(axis description cs:0.5,0.0)},anchor=north,font=\large},
                 y label style={at={(axis description cs:0.1,.5)},anchor=south,font=\large},
                 width=9cm,
                 height=7cm,
                 ymajorgrids=false,
                 xmajorgrids=false,
                 major grid style={dotted,green!20!black},
                 legend columns=2,
                 legend style={
                  nodes={scale=1.0, transform shape},
                  cells={anchor=west},
                  legend style={at={(0.36,0.0)},anchor=south,row sep=0.01pt}, font =\small}
             ]
             \addlegendentry{Dynamic}
             \addplot[line width=1pt, mark size=2pt, color=cyan, mark=triangle,error bars/.cd, y dir=both, y explicit,]
                     coordinates {(43.4, 51.44)
                                  (52.08, 49.69)
                                  (60.76, 47.69)
                                  (69.44, 42.77)
                                  (78.1, 37.46)
                                  (86.80, 28.05)};
            \addlegendentry{Static}
            \addplot[line width=1pt, mark size=2pt, color=orange, mark=square,error bars/.cd, y dir=both, y explicit,]
                     coordinates {(43.4, 53.436)
                                  (52.08, 52.469)
                                  (60.76, 51.293)
                                  (69.44, 49.62)
                                  (78.1, 46.587)
                                  (86.80, 39.799)};
            \addlegendentry{Random}
            \addplot[line width=1pt, mark size=2pt, color=green, mark=*,error bars/.cd, y dir=both, y explicit,]
                     coordinates {(43.4, 54.48)
                                  (52.08, 53.916)
                                  (60.76, 53.775)
                                  (69.44, 53.316)
                                  (78.1, 52.415)
                                  (86.80, 46.983)
                                  };
            \addlegendentry{Standard}
             \addplot[line width=1pt, mark size=0, color=gray, mark=diamond,error bars/.cd, y dir=both, y explicit,]
                     coordinates {(0, 54.9)
                                  (100, 54.9)};
              \end{axis}
\end{tikzpicture}
}
\end{center}
\caption{Top-$k$ unit removal experiments for the static and dynamic factors with respect to random units for action recognition. The final layer of the SlowFast model trained on the SSv2 dataset is considered.}\label{fig:ar_unit_removal}
\end{figure}

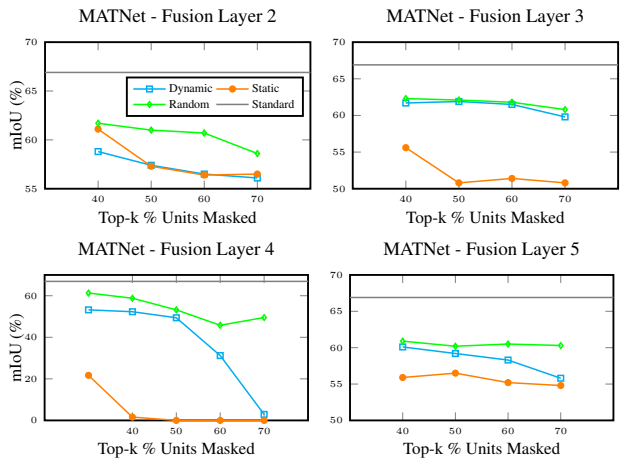
\begin{figure} [t]
	\begin{center}
     \centering 
\resizebox{0.48\textwidth}{!}{
\hspace{0.1cm}
\begin{tikzpicture} 
                 \begin{axis}[
                 line width=1.0,
                 title={MATNet - Fusion Layer 2},
                 title style={at={(axis description cs:0.5,1.2)},anchor=north,font=\normalsize},
                 xlabel={Top-k \% Units Masked},
                 ylabel={mIoU (\%)},
                 xmin=30, xmax=80,
                 ymin=55, ymax=70,
                 xtick={40,50,60,70},
                 x tick label style={font=\tiny, rotate=0, anchor=north},
                 y tick label style={font=\tiny},
                 x label style={at={(axis description cs:0.5,0.09)},anchor=north,font=\small},
                 y label style={at={(axis description cs:0.2,.5)},anchor=south,font=\small},
                 width=6.5cm,
                 height=4.3cm, 
                 ymajorgrids=false,
                 xmajorgrids=false,
                 major grid style={dotted,green!20!black},
                 legend columns=2,
                 legend style={
                  nodes={scale=0.7, transform shape},
                  cells={anchor=west},
                  legend style={at={(0.64,0.49)},anchor=south,row sep=0.01pt}, font =\small}
             ]

             \addlegendentry{Dynamic}
             \addplot[line width=0.8pt, mark size=1.5pt, color=cyan, mark=square,error bars/.cd, y dir=both, y explicit,]
                     coordinates {(40, 58.8)
                                  (50, 57.4)
                                  (60, 56.5)
                                  (70, 56.1)};
            \addlegendentry{Static}
            \addplot[line width=0.8pt, mark size=1.5pt, color=orange, mark=*,error bars/.cd, y dir=both, y explicit,]
                     coordinates {(40, 61.1)
                                  (50, 57.3)
                                  (60, 56.4)
                                  (70, 56.5)};
            \addlegendentry{Random}
            \addplot[line width=0.8pt, mark size=1.5pt, color=green, mark=diamond,error bars/.cd, y dir=both, y explicit,]
                     coordinates {(40, 61.7)
                                  (50, 61.0)
                                  (60, 60.7)
                                  (70, 58.6)
            };
            \addlegendentry{Standard}
             \addplot[line width=0.8pt, mark size=0, color=gray, mark=*,error bars/.cd, y dir=both, y explicit,]
                     coordinates {(30, 66.9)
                                  (80, 66.9)};
              \end{axis}
\end{tikzpicture}
\hspace{0.1cm}
\begin{tikzpicture} 
                 \begin{axis}[
                 line width=1.0,
                 title={MATNet - Fusion Layer 3},
                 title style={at={(axis description cs:0.5,1.2)},anchor=north,font=\normalsize},
                 xlabel={Top-k \% Units Masked},
                 ylabel={},
                 xmin=30, xmax=80,
                 ymin=50, ymax=70,
                 xtick={40,50,60,70},
                 x tick label style={font=\tiny, rotate=0, anchor=north},
                 y tick label style={font=\tiny},
                 x label style={at={(axis description cs:0.5,0.09)},anchor=north,font=\small},
                 y label style={at={(axis description cs:0.2,.5)},anchor=south,font=\small},
                 width=6.5cm,
                 height=4.3cm, 
                 ymajorgrids=false,
                 xmajorgrids=false,
                 major grid style={dotted,green!20!black},
             ]
            
             \addplot[line width=0.8pt, mark size=1.5pt, color=cyan, mark=square,error bars/.cd, y dir=both, y explicit,]
                     coordinates {(40, 61.7)
                                  (50, 61.9)
                                  (60, 61.5)
                                  (70, 59.8)};
            \addplot[line width=0.8pt, mark size=1.5pt, color=orange, mark=*,error bars/.cd, y dir=both, y explicit,]
                     coordinates {(40, 55.6)
                                  (50, 50.8)
                                  (60, 51.4)
                                  (70, 50.8)};
            \addplot[line width=0.8pt, mark size=1.5pt, color=green, mark=diamond,error bars/.cd, y dir=both, y explicit,]
                     coordinates {(40, 62.3)
                                  (50, 62.1)
                                  (60, 61.8)
                                  (70, 60.8)
            };
             \addplot[line width=0.8pt, mark size=0, color=gray, mark=*,error bars/.cd, y dir=both, y explicit,]
                     coordinates {(30, 66.9)
                                  (80, 66.9)};
              \end{axis}
\end{tikzpicture}}
\resizebox{0.48\textwidth}{!}{
\hspace{0.1cm}
\begin{tikzpicture} 
                 \begin{axis}[
                 line width=1.0,
                 title={MATNet - Fusion Layer 4},
                 title style={at={(axis description cs:0.5,1.2)},anchor=north,font=\normalsize},
                 xlabel={Top-k \% Units Masked},
                 ylabel={mIoU (\%)},
                 xmin=20, xmax=80,
                 ymin=0, ymax=70,
                 xtick={40,50,60,70},
                 x tick label style={font=\tiny, rotate=0, anchor=north},
                 y tick label style={font=\tiny},
                 x label style={at={(axis description cs:0.5,0.09)},anchor=north,font=\small},
                 y label style={at={(axis description cs:0.2,.5)},anchor=south,font=\small},
                 width=6.5cm,
                 height=4.3cm, 
                 ymajorgrids=false,
                 xmajorgrids=false,
                 major grid style={dotted,green!20!black},
             ]
            %
             \addplot[line width=0.8pt, mark size=1.5pt, color=cyan, mark=square,error bars/.cd, y dir=both, y explicit,]
                     coordinates {(30, 53.2)
                                  (40, 52.3)
                                  (50, 49.4)
                                  (60, 31.2)
                                  (70, 2.9)};
            \addplot[line width=0.8pt, mark size=1.5pt, color=orange, mark=*,error bars/.cd, y dir=both, y explicit,]
                     coordinates {(30, 21.7)
                                  (40, 1.6)
                                  (50, 1.969784267764861e-06)
                                  (60, 3.4545299701601246e-05)
                                  (70, 1.0245879243979795e-07)};
            \addplot[line width=0.8pt, mark size=1.5pt, color=green, mark=diamond,error bars/.cd, y dir=both, y explicit,]
                     coordinates {(30, 61.3)
                                  (40, 58.8)
                                  (50, 53.2)
                                  (60, 45.8)
                                  (70, 49.5)
            };
             \addplot[line width=0.8pt, mark size=0, color=gray, mark=*,error bars/.cd, y dir=both, y explicit,]
                     coordinates {(20, 66.9)
                                  (80, 66.9)};
              \end{axis}
\end{tikzpicture}
\hspace{0.1cm}
\begin{tikzpicture} 
                 \begin{axis}[
                 line width=1.0,
                 title={MATNet - Fusion Layer 5},
                 title style={at={(axis description cs:0.5,1.2)},anchor=north,font=\normalsize},
                 xlabel={Top-k \% Units Masked},
                 ylabel={},
                 xmin=30, xmax=80,
                 ymin=50, ymax=70,
                 xtick={40,50,60,70},
                 x tick label style={font=\tiny, rotate=0, anchor=north},
                 y tick label style={font=\tiny},
                 x label style={at={(axis description cs:0.5,0.09)},anchor=north,font=\small},
                 y label style={at={(axis description cs:0.2,.5)},anchor=south,font=\small},
                 width=6.5cm,
                 height=4.3cm, 
                 ymajorgrids=false,
                 xmajorgrids=false,
                 major grid style={dotted,green!20!black},
             ]
            
             \addplot[line width=0.8pt, mark size=1.5pt, color=cyan, mark=square,error bars/.cd, y dir=both, y explicit,]
                     coordinates {(40, 60.1)
                                  (50, 59.2)
                                  (60, 58.3)
                                  (70, 55.8)};
            \addplot[line width=0.8pt, mark size=1.5pt, color=orange, mark=*,error bars/.cd, y dir=both, y explicit,]
                     coordinates {(40, 55.9)
                                  (50, 56.5)
                                  (60, 55.2)
                                  (70, 54.8)};
            \addplot[line width=0.8pt, mark size=1.5pt, color=green, mark=diamond,error bars/.cd, y dir=both, y explicit,]
                     coordinates {(40, 60.9)
                                  (50, 60.2)
                                  (60, 60.5)
                                  (70, 60.3)
            };
             \addplot[line width=0.8pt, mark size=0, color=gray, mark=*,error bars/.cd, y dir=both, y explicit,]
                     coordinates {(30, 66.9)
                                  (80, 66.9)};
              \end{axis}
\end{tikzpicture}
}
\end{center}
\caption{Top-$k$ unit removal experiments for the static and dynamic factors with respect to random units for video object segmentation. The three fusion layers of the MATNet model trained on DAVIS and YouTube-VOS are considered.}\label{fig:vos_unit_removal}
\end{figure}

To evaluate the effect of the estimated static and dynamic biased units on overall performance, we conduct perturbation experiments. In these experiments, we remove the top $k$ units (\ie channels) that are biased toward the static or dynamic factor during inference and evaluate the final accuracy drop. The removal is done by setting all activations to zero in the identified channels. We compare these static or dynamic biased units with respect to randomly selected channels. Figures~\ref{fig:ar_unit_removal} and~\ref{fig:vos_unit_removal} show the unit removal results for the action recognition and video object segmentation experiments, respectively. In action recognition we remove the top-$k$ static, dynamic and random channels from the final layer in the SlowFast model trained on SSv2. We then evaluate on the SSv2 validation set and report the top-1 accuracy. As can be seen in Fig.~\ref{fig:ar_unit_removal}, the dynamic factor maximally reduces the model's performance, which may be because (i) the SlowFast model encodes a significant amount of dynamic information in the fast branch and (ii) dynamics are important to solve the SSv2 dataset. We conduct similar experiments on video object segmentation for the four fusion layers of the MATNet model trained on DAVIS and YouTube-VOS. We evaluate on the MoCA dataset and report the mean intersection over union (mIoU). The results in Fig.~\ref{fig:vos_unit_removal} consistently demonstrate that for every fusion layer the factor with the highest impact on performance is the factor it is most biased toward, as examined earlier in the main submission (Fig.~\ref{fig:stagewise_vos}).
In both tasks, these experiments document that masking out the top-$k$ channels based on our proposed static/dynamic bias estimate can help us control what the model is biased toward and consequently affect its accuracy compared with randomly selected channels.

\subsection{Computational load}\label{sec:comp_load}
We provide details of the models used in the paper in regards to their computational load. For each model, we list their FLOPs and parameter count in Table~\ref{tab:comp_load}. We do not observe any correlation between computational load and biases of the model and leave a deeper analysis of this connection for future work.

\begin{table}
\resizebox{0.45\textwidth}{!}{
\begin{tabu}{c cc}
\tabucline[1pt]{-}
 \multirow{2}{*}{Model} &\multicolumn{2}{c}{Action Recognition}  \\
\cline{2-3}
& Parameters (M) & GFLOPs \\
\tabucline[1pt]{-}
C2D & 24.3 & 25.6 \\
I3D & 28.0 & 37.3  \\
X3D-m & 3.8 &6.4 \\
SlowOnly & 32.5 & 54.8   \\
FastOnly & 0.6 &  7.0 \\
SlowFast (8x8) & 34.6 & 66.1 \\
MViT & 36.6 & 70.7 \\
TimeSformer & 121.6 &  196.1 \\ 
\tabucline[1pt]{-}
 \multirow{2}{*}{Model} &\multicolumn{2}{c}{Video Object Segmentation} \\
\cline{2-3}
& Parameters (M) & GFLOPs \\
\tabucline[1pt]{-}
MATNet & 142.7 & 156.0 \\
RTNet & 277.2 & 309.7 \\
FusionSeg & 113.0 & 112.5\\
\tabucline[1pt]x
\end{tabu}}
\caption{Computational loads of the action recognition and video object segmentation models studied.}
\label{tab:comp_load}
\end{table}

\subsection{Assets}\label{sec:assets}

\noindent \textbf{Action recognition.} We use provided code and trained weights from the SlowFast repository\footnote{\url{https://github.com/facebookresearch/SlowFast}} and TimeSformer repository\footnote{\url{https://github.com/facebookresearch/TimeSformer}}. SlowFast is licensed under the Apache 2.0 license\footnote{\url{https://github.com/facebookresearch/SlowFast/blob/main/LICENSE}}. TimeSformer is licensed under the CC-NC 4.0 International license\footnote{\url{https://github.com/facebookresearch/TimeSformer/blob/main/LICENSE}} and Apache 2.0 license\footnote{\url{https://github.com/facebookresearch/SlowFast/blob/main/LICENSE}}. We use the Kinetics-400\footnote{\url{https://github.com/cvdfoundation/kinetics-dataset}}, SSv2\footnote{\url{https://20bn.com/datasets/something-something}} and Diving48\footnote{\url{http://www.svcl.ucsd.edu/projects/resound/dataset.html}} datasets.

\noindent \textbf{Video object segmentation.} We use provided code and trained weights for MATNet\footnote{\url{https://github.com/tfzhou/MATNet}} and RTNet\footnote{\url{https://github.com/OliverRensu/RTNet}}. No accompanied licences are provided with the aforementioned code. Additionally, we use the DAVIS16\footnote{\url{https://davischallenge.org/davis2016/code.html}}, TAO-VOS\footnote{\url{http://www.vision.rwth-aachen.de/page/taovos}} and ImageNet VID\footnote{\url{http://vision.cs.utexas.edu/projects/fusionseg/training_data.html}} datasets.

\clearpage\newpage 

{\small
\bibliographystyle{ieee_fullname}
\bibliography{main}

\begin{thebibliography}{10}\itemsep=-1pt

\bibitem{aafaq2019video}
Nayyer Aafaq, Ajmal Mian, Wei Liu, Syed~Zulqarnain Gilani, and Mubarak Shah.
\newblock Video description: A survey of methods, datasets, and evaluation
  metrics.
\newblock {\em ACM Computing Surveys}, 52(6):1--37, 2019.

\bibitem{bertasius2021space}
Gedas Bertasius, Heng Wang, and Lorenzo Torresani.
\newblock Is space-time attention all you need for video understanding?
\newblock In {\em Proceedings of the International Conference on Machine
  Learning}, 2021.

\bibitem{buolamwini2018gender}
Joy Buolamwini and Timnit Gebru.
\newblock Gender shades: Intersectional accuracy disparities in commercial
  gender classification.
\newblock In {\em Proceedings of the Conference on Fairness, Accountability and
  Transparency}, pages 77--91, 2018.

\bibitem{carreira2017quo}
Joao Carreira and Andrew Zisserman.
\newblock Quo vadis, action recognition? {A} new model and the kinetics
  dataset.
\newblock In {\em Proceedings of the IEEE Conference on Computer Vision and
  Pattern Recognition}, pages 6299--6308, 2017.

\bibitem{chen2018encoder}
Liang-Chieh Chen, Yukun Zhu, George Papandreou, Florian Schroff, and Hartwig
  Adam.
\newblock Encoder-decoder with atrous separable convolution for semantic image
  segmentation.
\newblock In {\em Proceedings of the European Conference on Computer Vision},
  pages 801--818, 2018.

\bibitem{choi2019can}
Jinwoo Choi, Chen Gao, C.~E.~Joseph Messou, and Jia-Bin Huang.
\newblock Why can't {I} dance in the mall? {L}earning to mitigate scene bias in
  action recognition.
\newblock In {\em Proceedings of the Conference on Advances in Neural
  Information Processing Systems}, 2019.

\bibitem{dave2019towards}
Achal Dave, Pavel Tokmakov, and Deva Ramanan.
\newblock Towards segmenting anything that moves.
\newblock In {\em Proceedings of the IEEE/CVF International Conference on
  Computer Vision Workshops}, pages 0--0, 2019.

\bibitem{derpanis2012action}
Konstantinos~G. Derpanis, Mikhail Sizintsev, Kevin~J. Cannons, and Richard~P.
  Wildes.
\newblock Action spotting and recognition based on a spatiotemporal orientation
  analysis.
\newblock {\em IEEE Transactions on Pattern Analysis and Machine Intelligence},
  35(3):527--540, 2012.

\bibitem{derpanis2011spacetime}
Konstantinos~G. Derpanis and Richard~P. Wildes.
\newblock Spacetime texture representation and recognition based on a
  spatiotemporal orientation analysis.
\newblock {\em IEEE Transactions on Pattern Analysis and Machine Intelligence},
  34(6):1193--1205, 2011.

\bibitem{esser2020disentangling}
Patrick Esser, Robin Rombach, and Bjorn Ommer.
\newblock A disentangling invertible interpretation network for explaining
  latent representations.
\newblock In {\em Proceedings of the IEEE/CVF Conference on Computer Vision and
  Pattern Recognition}, pages 9223--9232, 2020.

\bibitem{caba2015activitynet}
Bernard~Ghanem Fabian Caba~Heilbron, Victor~Escorcia and Juan~Carlos Niebles.
\newblock Activity{N}et: A large-scale video benchmark for human activity
  understanding.
\newblock In {\em Proceedings of the IEEE Conference on Computer Vision and
  Pattern Recognition}, pages 961--970, 2015.

\bibitem{fan2021multiscale}
Haoqi Fan, Bo Xiong, Karttikeya Mangalam, Yanghao Li, Zhicheng Yan, Jitendra
  Malik, and Christoph Feichtenhofer.
\newblock Multiscale vision transformers.
\newblock {\em arXiv preprint arXiv:2104.11227}, 2021.

\bibitem{feichtenhofer2020x3d}
Christoph Feichtenhofer.
\newblock X3{D}: Expanding architectures for efficient video recognition.
\newblock In {\em Proceedings of the IEEE/CVF Conference on Computer Vision and
  Pattern Recognition}, pages 203--213, 2020.

\bibitem{feichtenhofer2019slowfast}
Christoph Feichtenhofer, Haoqi Fan, Jitendra Malik, and Kaiming He.
\newblock Slow{F}ast networks for video recognition.
\newblock In {\em Proceedings of the IEEE/CVF International Conference on
  Computer Vision}, pages 6202--6211, 2019.

\bibitem{feichtenhofer2017spatiotemporal}
Christoph Feichtenhofer, Axel Pinz, and Richard~P Wildes.
\newblock Spatiotemporal multiplier networks for video action recognition.
\newblock In {\em Proceedings of the IEEE Conference on Computer Vision and
  Pattern Recognition}, pages 4768--4777, 2017.

\bibitem{feichtenhofer2020deep}
Christoph Feichtenhofer, Axel Pinz, Richard~P Wildes, and Andrew Zisserman.
\newblock Deep insights into convolutional networks for video recognition.
\newblock {\em International Journal of Computer Vision}, 128(2):420--437,
  2020.

\bibitem{foster2011lower}
David~V Foster and Peter Grassberger.
\newblock Lower bounds on mutual information.
\newblock {\em Physical Review E}, 83(1):010101, 2011.

\bibitem{ghodrati2018video}
Amir Ghodrati, Efstratios Gavves, and Cees G.~M. Snoek.
\newblock Video time: Properties, encoders and evaluation.
\newblock In {\em Proceedings of the British Machine Vision Conference}, 2018.

\bibitem{goyal2017something}
Raghav Goyal, Samira Ebrahimi~Kahou, Vincent Michalski, Joanna Materzynska,
  Susanne Westphal, Heuna Kim, Valentin Haenel, Ingo Fruend, Peter Yianilos,
  Moritz Mueller-Freitag, Florian Hoppe, Christian Thurau, Ingo Bax, and Roland
  Memisevic.
\newblock The ``something something" video database for learning and evaluating
  visual common sense.
\newblock In {\em Proceedings of the IEEE International Conference on Computer
  Vision}, pages 5842--5850, 2017.

\bibitem{hadji2018new}
Isma Hadji and Richard~P Wildes.
\newblock A new large scale dynamic texture dataset with application to convnet
  understanding.
\newblock In {\em Proceedings of the European Conference on Computer Vision},
  pages 320--335, 2018.

\bibitem{hansson2021self}
Sven~Ove Hansson, Matts-{\AA}ke Belin, and Bj{\"o}rn Lundgren.
\newblock Self-driving vehicles-{A}n ethical overview.
\newblock {\em Philosophy \& Technology}, pages 1--26, 2021.

\bibitem{hara2017learning}
Kensho Hara, Hirokatsu Kataoka, and Yutaka Satoh.
\newblock Learning spatio-temporal features with 3{D} residual networks for
  action recognition.
\newblock In {\em Proceedings of the IEEE International Conference on Computer
  Vision Workshops}, pages 3154--3160, 2017.

\bibitem{he2016deep}
Kaiming He, Xiangyu Zhang, Shaoqing Ren, and Jian Sun.
\newblock Deep residual learning for image recognition.
\newblock In {\em Proceedings of the IEEE Conference on Computer Vision and
  Pattern Recognition}, pages 770--778, 2016.

\bibitem{he2016human}
Yun He, Soma Shirakabe, Yutaka Satoh, and Hirokatsu Kataoka.
\newblock Human action recognition without human.
\newblock In {\em Proceedings of the European Conference on Computer Vision},
  pages 11--17, 2016.

\bibitem{hiley2019explainable}
Liam Hiley, Alun Preece, and Yulia Hicks.
\newblock Explainable deep learning for video recognition tasks: A framework \&
  recommendations.
\newblock {\em arXiv preprint arXiv:1909.05667}, 2019.

\bibitem{huang2017arbitrary}
Xun Huang and Serge Belongie.
\newblock Arbitrary style transfer in real-time with adaptive instance
  normalization.
\newblock In {\em Proceedings of the IEEE International Conference on Computer
  Vision}, pages 1501--1510, 2017.

\bibitem{islam2021shape}
Md~Amirul Islam, Matthew Kowal, Patrick Esser, Sen Jia, Bj{\"o}rn Ommer,
  Konstantinos~G. Derpanis, and Neil D.~B. Bruce.
\newblock Shape or texture: Understanding discriminative features in
  {C}{N}{N}s.
\newblock In {\em Proceedings of the International Conference on Learning
  Representations}, 2021.

\bibitem{jain2017fusionseg}
Suyog~Dutt Jain, Bo Xiong, and Kristen Grauman.
\newblock Fusion{S}eg: Learning to combine motion and appearance for fully
  automatic segmentation of generic objects in videos.
\newblock In {\em Proceedings of the IEEE Conference on Computer Vision and
  Pattern Recognition}, pages 2117--2126. IEEE, 2017.

\bibitem{ji20123d}
Shuiwang Ji, Wei Xu, Ming Yang, and Kai Yu.
\newblock 3{D} convolutional neural networks for human action recognition.
\newblock {\em IEEE Transactions on Pattern Analysis and Machine Intelligence},
  35(1):221--231, 2012.

\bibitem{kraskov2004estimating}
Alexander Kraskov, Harald St{\"o}gbauer, and Peter Grassberger.
\newblock Estimating mutual information.
\newblock {\em Physical review E}, 69(6):066138, 2004.

\bibitem{lamdouar2020betrayed}
Hala Lamdouar, Charig Yang, Weidi Xie, and Andrew Zisserman.
\newblock Betrayed by motion: Camouflaged object discovery via motion
  segmentation.
\newblock In {\em Proceedings of the Asian Conference on Computer Vision},
  2020.

\bibitem{d48_web}
Yingwei Li, Yi Li, and Nuno Vasconcelos.
\newblock Diving48 dataset.
\newblock \url{http://www.svcl.ucsd.edu/projects/resound/dataset.html}.
\newblock Accessed: 2021-11-13.

\bibitem{li2018resound}
Yingwei Li, Yi Li, and Nuno Vasconcelos.
\newblock Resound: Towards action recognition without representation bias.
\newblock In {\em Proceedings of the European Conference on Computer Vision},
  pages 513--528, 2018.

\bibitem{li2019repair}
Yi Li and Nuno Vasconcelos.
\newblock Repair: Removing representation bias by dataset resampling.
\newblock In {\em Proceedings of the IEEE/CVF Conference on Computer Vision and
  Pattern Recognition}, pages 9572--9581, 2019.

\bibitem{manttari2020interpreting}
Joonatan Manttari, Sofia Broom{\'e}, John Folkesson, and Hedvig Kjellstrom.
\newblock Interpreting video features: {A} comparison of 3{D} convolutional
  networks and convolutional {L}{S}{T}{M} networks.
\newblock In {\em Proceedings of the Asian Conference on Computer Vision},
  2020.

\bibitem{patrick2021keeping}
Mandela Patrick, Dylan Campbell, Yuki~M. Asano, Ishan Misra~Florian Metze,
  Christoph Feichtenhofer, Andrea Vedaldi, and Joao~F. Henriques.
\newblock Keeping your eye on the ball: Trajectory attention in video
  transformers.
\newblock In {\em Proceedings of the Conference on Advances in Neural
  Information Processing Systems}, 2021.

\bibitem{Perazzi2016}
F. Perazzi, J. Pont-Tuset, B. McWilliams, L. {Van Gool}, M. Gross, and A.
  Sorkine-Hornung.
\newblock A benchmark dataset and evaluation methodology for video object
  segmentation.
\newblock In {\em Proceedings of the IEEE Conference on Computer Vision and
  Pattern Recognition}, 2016.

\bibitem{ren2021reciprocal}
Sucheng Ren, Wenxi Liu, Yongtuo Liu, Haoxin Chen, Guoqiang Han, and Shengfeng
  He.
\newblock Reciprocal transformations for unsupervised video object
  segmentation.
\newblock In {\em Proceedings of the IEEE/CVF Conference on Computer Vision and
  Pattern Recognition}, pages 15455--15464, 2021.

\bibitem{sevilla2021only}
Laura Sevilla-Lara, Shengxin Zha, Zhicheng Yan, Vedanuj Goswami, Matt Feiszli,
  and Lorenzo Torresani.
\newblock Only time can tell: Discovering temporal data for temporal modeling.
\newblock In {\em Proceedings of the IEEE/CVF Winter Conference on Applications
  of Computer Vision}, pages 535--544, 2021.

\bibitem{simonyan2014two}
Karen Simonyan and Andrew Zisserman.
\newblock Two-stream convolutional networks for action recognition in videos.
\newblock In {\em Proceedings of the Conference on Advances in Neural
  Information Processing Systems}, volume~27, 2014.

\bibitem{taylor2010convolutional}
Graham~W Taylor, Rob Fergus, Yann LeCun, and Christoph Bregler.
\newblock Convolutional learning of spatio-temporal features.
\newblock In {\em Proceedings of the European Conference on Computer Vision},
  pages 140--153, 2010.

\bibitem{teed2020raft}
Zachary Teed and Jia Deng.
\newblock Raft: Recurrent all-pairs field transforms for optical flow.
\newblock In {\em Proceedings of the European Conference on Computer Vision},
  pages 402--419. Springer, 2020.

\bibitem{texler2020interactive}
Ond{\v{r}}ej Texler, David Futschik, Michal Ku{\v{c}}era, Ond{\v{r}}ej
  Jamri{\v{s}}ka, {\v{S}}{\'a}rka Sochorov{\'a}, Menclei Chai, Sergey Tulyakov,
  and Daniel S{\`y}kora.
\newblock Interactive video stylization using few-shot patch-based training.
\newblock {\em ACM Transactions on Graphics (TOG)}, 39(4):73--1, 2020.

\bibitem{tokmakov2017learning}
Pavel Tokmakov, Karteek Alahari, and Cordelia Schmid.
\newblock Learning video object segmentation with visual memory.
\newblock In {\em Proceedings of the IEEE International Conference on Computer
  Vision}, pages 4481--4490, 2017.

\bibitem{tran2015learning}
Du Tran, Lubomir Bourdev, Rob Fergus, Lorenzo Torresani, and Manohar Paluri.
\newblock Learning spatiotemporal features with 3{D} convolutional networks.
\newblock In {\em Proceedings of the IEEE International Conference on Computer
  Vision}, pages 4489--4497, 2015.

\bibitem{vaswani2017attention}
Ashish Vaswani, Noam Shazeer, Niki Parmar, Jakob Uszkoreit, Llion Jones,
  Aidan~N Gomez, {\L}ukasz Kaiser, and Illia Polosukhin.
\newblock Attention is all you need.
\newblock In {\em Proceedings of the Conference on Advances in Neural
  Information Processing Systems}, pages 5998--6008, 2017.

\bibitem{Voigtlaender21WACV}
Paul Voigtlaender, Lishu Luo, Chun Yuan, Yong Jiang, and Bastian Leibe.
\newblock Reducing the annotation effort for video object segmentation
  datasets.
\newblock In {\em Proceedings of the IEEE Winter Conference on Computer Vision
  Applications}, 2021.

\bibitem{vu2014predicting}
Tuan-Hung Vu, Catherine Olsson, Ivan Laptev, Aude Oliva, and Josef Sivic.
\newblock Predicting actions from static scenes.
\newblock In {\em Proceedings of the European Conference on Computer Vision},
  pages 421--436, 2014.

\bibitem{wang2021survey}
Wenguan Wang, Tianfei Zhou, Fatih Porikli, David Crandall, and Luc Van~Gool.
\newblock A survey on deep learning technique for video segmentation.
\newblock {\em arXiv preprint arXiv:2107.01153}, 2021.

\bibitem{wang2018non}
Xiaolong Wang, Ross Girshick, Abhinav Gupta, and Kaiming He.
\newblock Non-local neural networks.
\newblock In {\em Proceedings of the IEEE Conference on Computer Vision and
  Pattern Recognition}, pages 7794--7803, 2018.

\bibitem{wildes2000qualitative}
Richard~P Wildes and James~R Bergen.
\newblock Qualitative spatiotemporal analysis using an oriented energy
  representation.
\newblock In {\em European Conference on Computer Vision}, pages 768--784.
  Springer, 2000.

\bibitem{xu2018youtube}
Ning Xu, Linjie Yang, Yuchen Fan, Dingcheng Yue, Yuchen Liang, Jianchao Yang,
  and Thomas Huang.
\newblock You{T}ube-{VOS}: A large-scale video object segmentation benchmark.
\newblock {\em arXiv preprint arXiv:1809.03327}, 2018.

\bibitem{yang2021selfsupervised}
Charig Yang, Hala Lamdouar, Erika Lu, Andrew Zisserman, and Weidi Xie.
\newblock Self-supervised video object segmentation by motion grouping.
\newblock In {\em Proceedings of the International Conference on Computer
  Vision}, 2021.

\bibitem{zhao2021interpretable}
He Zhao and Richard~P Wildes.
\newblock Interpretable deep feature propagation for early action recognition.
\newblock {\em arXiv preprint arXiv:2107.05122}, 2021.

\bibitem{zhou2020motion}
Tianfei Zhou, Shunzhou Wang, Yi Zhou, Yazhou Yao, Jianwu Li, and Ling Shao.
\newblock Motion-attentive transition for zero-shot video object segmentation.
\newblock In {\em Proceedings of the AAAI Conference on Artificial
  Intelligence}, pages 13066--13073, 2020.

\bibitem{zhu2020comprehensive}
Yi Zhu, Xinyu Li, Chunhui Liu, Mohammadreza Zolfaghari, Yuanjun Xiong, Chongruo
  Wu, Zhi Zhang, Joseph Tighe, R Manmatha, and Mu Li.
\newblock A comprehensive study of deep video action recognition.
\newblock {\em arXiv preprint arXiv:2012.06567}, 2020.

\end{thebibliography}
}




\end{document}